\documentclass[lettersize,journal]{IEEEtran}
\usepackage{amsmath,amsfonts}
\usepackage{algorithm}
\usepackage{array}
\usepackage[caption=false,font=footnotesize,labelfont=rm,textfont=rm]{subfig}
\usepackage{textcomp}
\usepackage{stfloats}
\usepackage{url}
\usepackage{verbatim}
\usepackage{graphicx}
\usepackage{cite}
\usepackage{subcaption}
\usepackage{subfig}
\usepackage[noend]{algorithmic}
\usepackage{xcolor}   
\usepackage{multirow}
\begin{document}

\title{IS-Diff: Improving Diffusion-Based Inpainting with Better Initial Seed}

\author{Yongzhe Lyu,
        Yu Wu,
        Yutian Lin,
        and Bo Du%
\thanks{(Corresponding author: Yutian Lin and Bo Du.)}%
\thanks{The authors are with the
 School of Computer Science, Wuhan University, Wuhan 430072, China
 (e-mail: yongzhelv@whu.edu.cn; yutian.lin@whu.edu.cn; wuyucs@whu.edu.cn; dubo@whu.edu.cn)}%

}

\markboth{Journal of \LaTeX\ Class Files,~Vol.~14, No.~8, August~2021}%
{Shell \MakeLowercase{\textit{et al.}}: A Sample Article Using IEEEtran.cls for IEEE Journals}


\maketitle

\begin{abstract}
 {Diffusion models have shown promising results in free-form inpainting. Recent studies based on refined diffusion samplers or novel architectural designs have produced realistic results with improved contextual coherence between the inpainted regions and the visible image content. However, the random initial noise/state adopted in the vanilla diffusion process may introduce mismatched semantic information in masked regions, leading to biased inpainting results, e.g., low semantic coherence and inconsistent appearance with the unmasked regions.} 
 To address this issue, we propose the Initial Seed refined Diffusion Model (IS-Diff), a completely training-free approach incorporating distributional harmonious seeds to produce harmonious results. {Specifically, IS-Diff constructs a primary seed initialization by sampling from the unmasked regions to approximate the masked-region data distribution, and then combines it with random noise to form a compatible initial noise/state, thereby setting a promising direction for the inpainting.} Moreover, a dynamic selective refinement mechanism is proposed to detect severe unharmonious inpaintings in intermediate latent and adjust the strength of our initialization prior dynamically. We validate our method on both standard and large-mask inpainting tasks using the CelebA-HQ, ImageNet, and Places2 datasets, demonstrating its effectiveness across all metrics compared to state-of-the-art inpainting methods.

 \section{Introduction}
 
\end{abstract}

\IEEEPARstart{F}{ree-form} image inpainting, aiming at filling masked regions within an image without user-defined guidance, is one of the fundamental tasks in computer vision with various practical applications~\cite{Wang_Saharia_Montgomery_Pont_Tuset_Noy_Pellegrini_Onoe_Laszlo_Fleet_Soricut_et_al._2022, Zhang_Chang_2021, Vaquero_Turk_Pulli_Tico_Gelfand_2010}.
Despite the promising inpainting performance of generative adversarial networks (GANs)~\cite{Goodfellow_Pouget_Abadie_Mirza_Xu_Warde_Farley_Ozair_Courville_Bengio_2017} and autoregressive approaches~\cite{Wan_Zhang_Chen_Liao_2021, Yu_Zhan_WU_Pan_Cui_Lu_Ma_Xie_Miao_2021}, their dependence on training with specific datasets limits their ability to generalize effectively to unseen masks~\cite{Lugmayr_Danelljan_Romero_Yu_Timofte_Van_Gool_2022}.
\begin{figure}[htb]
    \centering  
    \subfloat[Previous methods]{%
\includegraphics[width=0.48\textwidth]{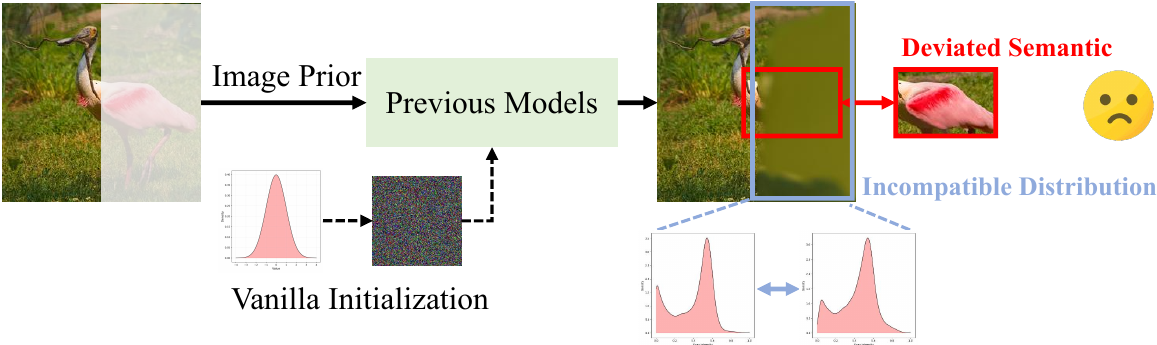}%
\label{fig:challenge_a}%
}
    \hfill
    \subfloat[Our proposed IS-Diff method]{%
\includegraphics[width=0.48\textwidth]{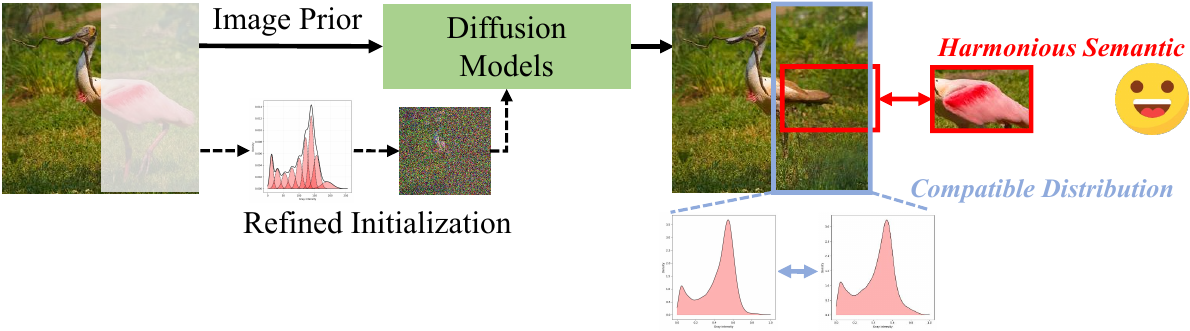}%
\label{fig:challenge_b}%
}
    \caption{Existing issues with current diffusion-based models and our proposed training-free method to tackle these problems. We use grayscale distribution maps to demonstrate the distributional discrepancy between inpainting result and reference. (a) Contemporary free-form inpainting techniques (DDNM~\cite{Wang_Yu_Zhang_2022} in this figure) often fail to create convincing semantic content and maintain cohesive textures because of distributional discrepancy between random initialization and the source image. (b) IS-Diff mitigates these challenges by utilizing information from the source image to close the distribution gap.}
    \label{challenge}
\end{figure}
\begin{figure*}[!t] 
		\centering 
         
	\includegraphics[width=0.9\textwidth]{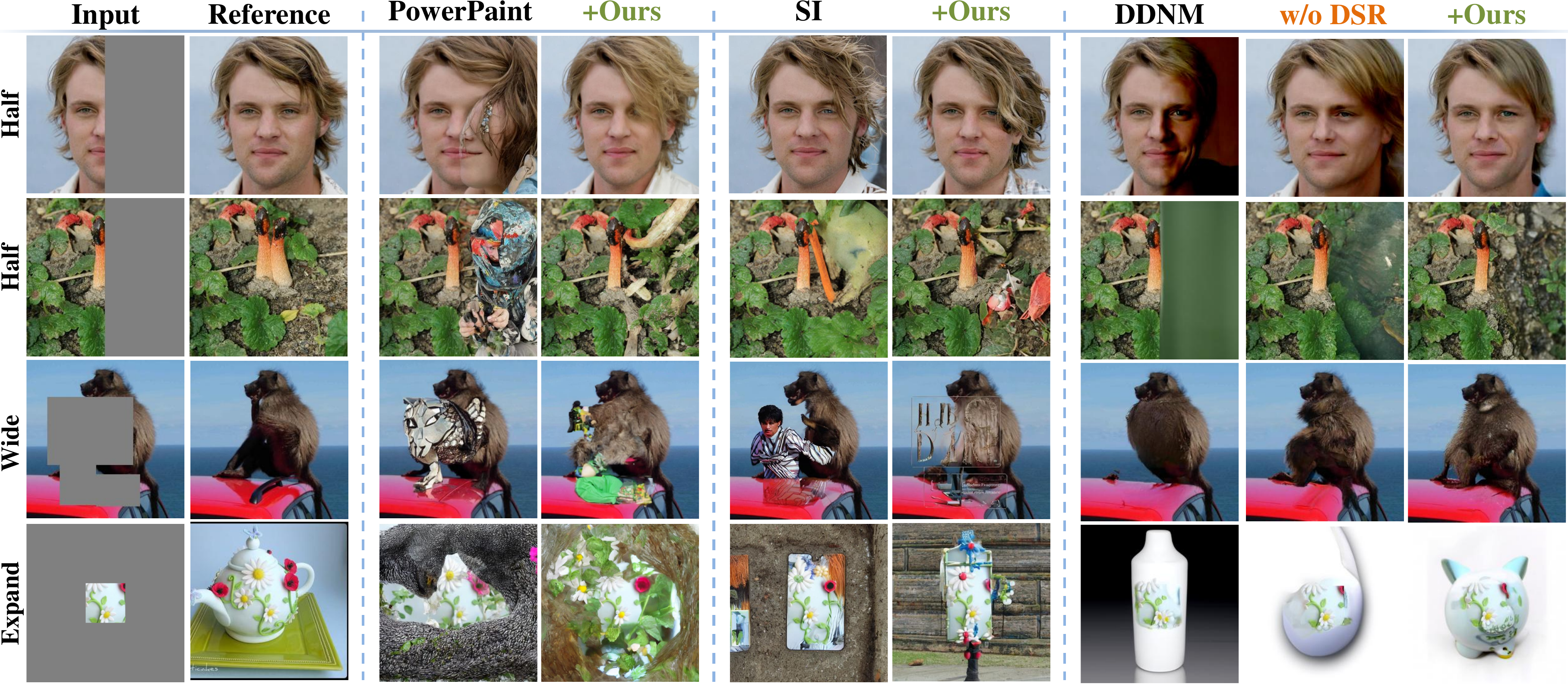} 
	\caption{Qualitative examples of different inpainting methods from the ImageNet 1K~\cite{Russakovsky_Deng_Su_Krause_Satheesh_Ma_Huang_Karpathy_Khosla_Bernstein_et_al._2015} dataset and the CelebA 1K~\cite{Liu_Luo_Wang_Tang_2015} dataset. ``DSR'' denotes our dynamic selective refinement, ``PP'' denotes PowerPaint~\cite{zhuang2023task}, and ``SI'' denotes {Stable Diffusion Inpainting}~\cite{Rombach_Blattmann_Lorenz_Esser_Ommer_2022}. Compared with various diffusion-based baselines, our plug-and-play method achieves more harmonious results on all types of masks.}

		\label{fig:intro} 
	\end{figure*}

Diffusion models~\cite{Dhariwal_Nichol_2021,Ho_Jain_Abbeel_2020,Kawar_Elad_Ermon_Song} have recently made remarkable progress, demonstrating robust semantic understanding and generation capabilities. Consequently, researches have surged in leveraging prior knowledge from pre-trained diffusion models for image inpainting~\cite{Lugmayr_Danelljan_Romero_Yu_Timofte_Van_Gool_2022, Wang_Yu_Zhang_2022, Liu_2024_CVPR, Luo_Gustafsson_Zhao_Sj_olund_Sch_on_2023}. These models have shown superior generalization capability and realistic visual effects, thanks to the powerful generative capabilities of diffusion models. 

However, as shown in Fig. 1 (a), the aforementioned diffusion-based methods still struggle to produce realistic inpainting results. In particular, they often fail to maintain semantic coherence, such as generating structures inconsistent with the original scene or introducing irrelevant content, and they also suffer from low-level inconsistencies, including mismatched colors, unnatural boundaries, and discontinuous textures. 
In this work, we attribute these deficiencies to an intrinsic conflict between incompatible vanilla Gaussian initialization and prescribed image priors. 
We conduct series of analyses to further explore this motivation in Sec.~\ref{moti}. In the context of free-form inpainting, the given image prior is often used to replace the masked regions of intermediate latents in diffusion to achieve better consistency while the unmasked regions are generated from pure noise.  Recent studies have observed that  an incompatible initialization can introduce conflict semantic content and obstructing high-quality image synthesis~\cite{guo2024initno, chen2024tino, zhou2024golden}. 
We conjecture that employing initializations with similar distributions facilitates better harmonization between the masked and unmasked regions of the image, leading to improved inpainting results.

Leveraging this insight, we propose IS-Diff, a training-free method based on enhanced diffusion initialization. To address the conflict issue mentioned above, instead of using vanilla initialization drawn from a Gaussian distribution, we propose to sample the seed from unmasked image areas to mimic the whole data distribution (as shown in Fig.~\ref{challenge} (b)). This strategy helps prevent the generation from irrational inpaintings of previous works, such as filling masked regions with a uniform color. However, during the diffusion process,
random noise is added to the refined initialization, which can
lead to extra uncertainty.
To solve that issue, we dynamically adjust the ratio of initialization to noise to select a harmonious initialization. Specifically, we propose the distributional cross-entropy (DCE) to evaluate diffusion intermediate latent. When the DCE value exceeds the threshold, which indicates severe misalignment between the masked and unmasked regions, we further refine the inpainting results by increasing the ratio of initialization to noise, enabling better initialization selection and harmonious inpainting.
It is worth noting that the improved initial seeds also incorporate randomness, resulting in outcomes that are both high-quality and highly diverse. We conduct quantitative and qualitative experiments to discuss the diversity of our method in~\ref{4.4}.

More results are illustrated in Sec.~\ref{sec:Experiment}, where extensive experiments conducted on ImageNet, CelebA-HQ and Places2 with both standard and large masks demonstrate our superiority in both visual quality and quantitive performance compared to state-of-the-art methods. Notably, our method is plug-and-play, and can be seamlessly integrated into diffusion-based pipelines like RePaint~\cite{Lugmayr_Danelljan_Romero_Yu_Timofte_Van_Gool_2022}, DDNM~\cite{Wang_Yu_Zhang_2022} and {Stable Diffusion Inpainting}~\cite{Rombach_Blattmann_Lorenz_Esser_Ommer_2022}, consistently yielding performance gains without any additional implementation or computational overhead.
 Our contributions can be summarized as follows:
\begin{itemize}
\item {We identify that distributionally compatible initialization is crucial to satisfactory free-form inpainting.}
\item We propose IS-Diff, where a semantically meaningful initial seed is constructed to drive the diffusion process, leading to coherent and consistent inpainting. 
\item We introduce {a} dynamic {selective} refinement mechanism

\end{itemize} 

\begin{figure*}[!t] 
		\centering 

            \includegraphics[width=0.9\linewidth]{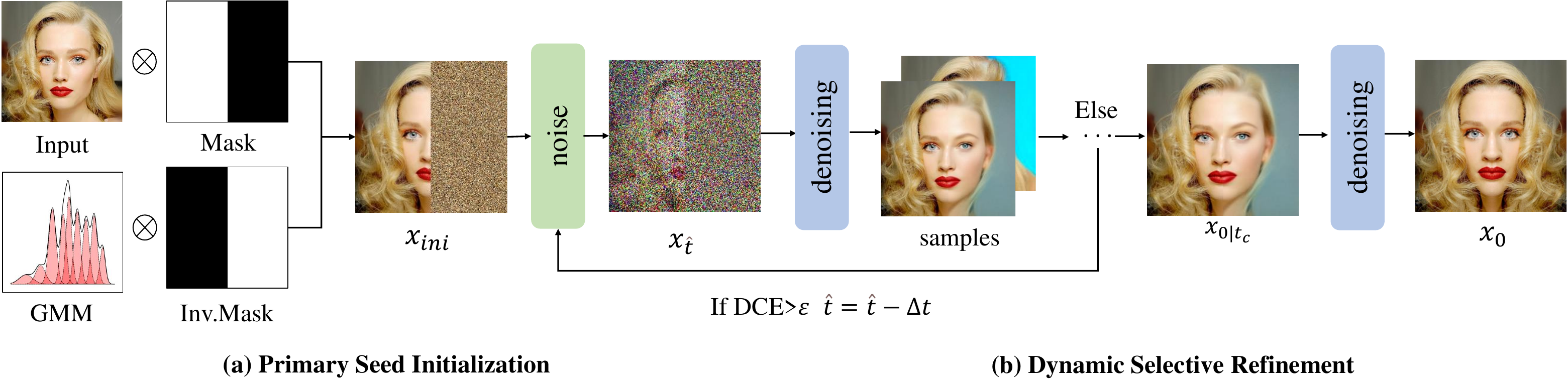}
            \caption{Overview of our approach. (a) for the primary seed initialization, masked regions are initialized with unmasked data distribution. (b) we calculate distribution cross-entropy of the intermediate variables of diffusion to implement selective iterative generation, and adjust the initialization strength through feedback, resulting in efficient and powerful inpainting.}
		\label{method} 
\end{figure*}
\section{Related works}
\label{gen_inst}

\subsection{Image Inpainting}
GAN-based methods with adversarial training have achieved high performance in standard and deterministic image inpainting tasks by introducing manually designed network architectures and loss functions to guide the generator~\cite{Iizuka_Simo-Serra_Ishikawa_2017,Nazeri_Ng_Joseph_Qureshi_Ebrahimi_2019,Xiong_Yu_Lin_Yang_Lu_Barnes_Luo_2019,Yu_Koltun_2015,Song_Yang_Lin_Liu_Huang_Li_Kuo_2018,Han_Wu_Wu_Yu_Davis_2018}. Autoregressive-based methods introduce various architectures to help effective generation, including Partial Convolutions~\cite{Liu_Reda_Shih_Wang_Tao_Catanzaro_2018}, Gated Convolutions~\cite{Yu_Lin_Yang_Shen_Lu_Huang_2019}, and Contextual Attention~\cite{Yu_Lin_Yang_Shen_Lu_Huang_2018}. Recently, diverse image inpainting methods have developed. As an ill-posed task, the diversity of generation is crucial for image inpainting. To generate pluralistic results, Zhao \textit{et al}.~\cite{Zhao_Mo_Lin_Wang_Zuo_Chen_Xing_Lu_2020} and Han \textit{et al}.~\cite{Han_Wu_Huang_Scott_Davis_2019} train VAE~\cite{Walker_Doersch_Gupta_Hebert_2016} type of networks to stochastically sample from Gaussian distribution conditioned by masked images. Liu \textit{et al}.~\cite{Liu_Wan_Huang_Song_Han_Liao_2021} achieve diverse generation by sampling a random seed from standard Gaussian distribution and mapping the seed into the image by a single decoder. However, all these methods rely on training on specific datasets. Moreover, GAN-based methods are usually difficult to train, while autoregressive methods tend to generate more average images, lacking diversity in generation.

\subsection{Diffusion-Based Image Inpainting}
SohlDickstein \textit{et al.}~\cite{Sohl-Dickstein_Weiss_Maheswaranathan_Ganguli_2015} apply early diffusion models to inpainting. Song \textit{et al.}~\cite{Song_Sohl-Dickstein_Kingma_Kumar_Ermon_Poole_2020} and Luo \textit{et al.}~\cite{Luo_Gustafsson_Zhao_Sj_olund_Sch_on_2023} achieve a range of applications based on stochastic differential equations, including image inpainting. Lugmayr \textit{et al}.~\cite{Lugmayr_Danelljan_Romero_Yu_Timofte_Van_Gool_2022} show a simple way to utilize image prior based on pretrained Diffusion Models by constantly adding the unmasked part of masked images to guide the model to generate the missing regions. RePaint~\cite{Lugmayr_Danelljan_Romero_Yu_Timofte_Van_Gool_2022} also proposes resampling to harmonize the inpainting result by reversing $x_t$ to an earlier timestep and resampling repeatedly. RePaint outperforms previous GAN based methods in terms of realism and performance on large masks. Wang \textit{et al}.~\cite{Wang_Yu_Zhang_2022} propose DDNM to guide diffusion model to solve image restoration tasks based on range-null space decomposition by constantly replacing the range-space part of $x_{0 \mid t}$ with that of degraded image y. 
Recent methods based on latent diffusion model~\cite{Zhuang_Zeng_Liu_Yuan_Chen_2023, Ju2024BrushNetAP} improve the quality of text-guided inpainting by novelly fine-tuning pre-trained models. Manukyan \textit{et al.}~\cite{Manukyan_Sargsyan_Atanyan_Wang_Navasardyan_Shi} introduce a training-free attention layer to better align the inpainted area with user prompts and Liu \textit{et al.}~\cite{Liu_2024_CVPR} design a structure guidance diffusion model to improve data consistency for inpainting. 

\subsection{Noise Optimization}
Recent work has revealed that the random Gaussian noise used to initialize diffusion samplers is not merely a stochastic nuisance but an influential control variable: different initial noises can lead to systematically different image quality, layout, and prompt-alignment outcomes~\cite{zhou2024golden}. Thus, methods optimize initial noise based on selection or optimization have emerged recently~\cite{zhou2024golden, guo2024initno, eyring2024reno, Zhang2025InferencetimeSO, qi2024not}. Selection methods pick seeds with favorable evaluation functions, while optimization methods directly alter the noise. For optimization-based methods, InitNO steers latents using attention signals~\cite{guo2024initno}, while ReNO optimizes noise with reward signals~\cite{eyring2024reno}, and Golden Noise proposes a lightweight learning framework to transform random noise into improved ``golden noise"~\cite{zhou2024golden}. For selection-based methods, Ma \textit{et al.}~\cite{Zhang2025InferencetimeSO} designs a mechanism to select initializations based on specific evaluation metrics. To the best of our knowledge, we are the first to apply noise optimization in free-form inpainting, advancing the field by focusing on enhancing both the inpainting quality and consistency.


\begin{figure*}[htb] 
		\centering 

            \includegraphics[width=0.9\linewidth]{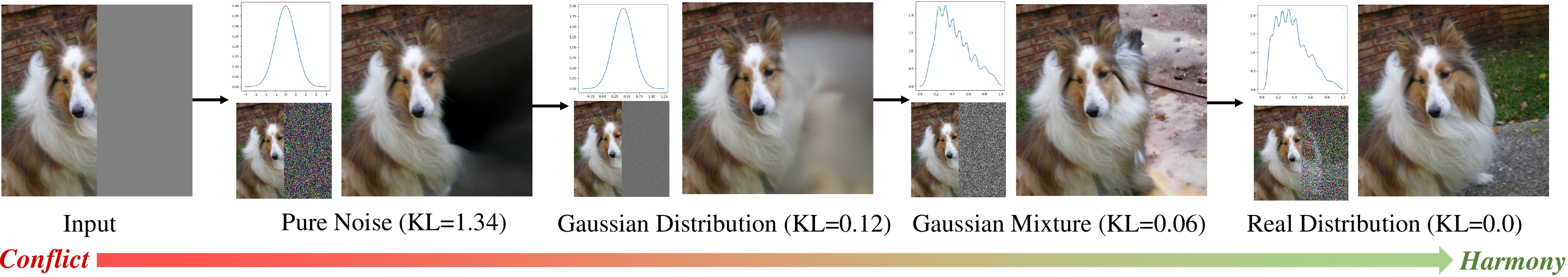}

                        \caption{We analyze different initializations to explore how distribution differences inside and outside the mask affect inpainting results.  ``KL'' denotes the KL divergence between the initialization distribution and the real distribution. The results indicate that initializations with similar distributions have a positive impact on inpainting. When initialized with pure noise, the generated contents tend to be hollow and lack structural consistency. As the KL divergence decreases, the inpainting results gradually become clearer, with the main subject better aligned and more harmonious. In particular, when directly sampling from the distribution of the unmasked regions, the inpainting results reach the highest quality.}
		\label{motivation} 
\end{figure*}
\section{Method}
\subsection{Better Initialization For Diffusion Inpainting} 
\label{moti}
Diffusion Models (DMs) define a $T$-step forward process and a $T$-step reverse process. Random noise is added gradually to data while the reverse process is to reconstruct data from pure Gaussian noise. The forward process yields $x_t$ from $x_{t-1}$ as below:
\begin{equation}
x_{t}=\sqrt{1-\beta_t}x_{t-1}+\sqrt{\beta_t}\varepsilon, \label{DDPM forward}
\end{equation}
where, $\beta_t \in (0,1]$ is the variance schedule, and $\varepsilon$ denotes the Gaussian noise. 
The diffusion model is trained to reverse the forward process in Eq.~(\ref{DDPM forward}).
Diffusion models uses a neural network to predict residual noise and parameters $\mu_\theta(x_t,t)$ and $\Sigma_\theta(x_t,t)$ of a Gaussian distribution:
\begin{equation}
p_\theta(x_{t-1}\mid x_{t})=N(x_{t-1};\mu_\theta(x_t,t),\Sigma_\theta(x_t,t)). \label{reverse}
\end{equation}
Since the add-noise step in Eq.~(\ref{DDPM forward}) is finite, the signal-to-noise ratio of $x_t$ during the forward process keeps positive, so that the start code $x_T$ has an impact on inpainting results.
In the context of free-form inpainting, given the original image $x\in\mathbb{R}^{H\times W \times 3}$ and the binary mask $m\in\{0,1\}^{H,W}$, the observed corrupted image $y$ is defined as $y=x\odot m$. The inpainting process can be represented as:
\begin{equation}
\hat{x} = (1-m) \odot f_\theta (y,m,ini) + m\odot y ,
\label{inpainting}
\end{equation}
where $\hat{x}$ denotes inpainting result, $f_\theta$ denotes diffusion model and $ini$ denotes the initialization of diffusion process, which is sampled from gaussian noise in vanilla diffusion. The given image $y$ and initialization $ini$ jointly influence the generated results, making the compatibility between the two even more vital for coherent inpainting.

In previous diffusion-based inpainting methods, models are guided to generate inpainting results by starting from random noise and continuously replacing corresponding regions of intermediate variables of the diffusion process with known parts. However, this approach often leads to inconsistencies between the newly generated and unmasked regions and may result in failed inpainting outcomes when the masked region is large. 

{The issue primarily arises from a fundamental distributional mismatch at the beginning of the reverse process. To formalize why vanilla Gaussian noise struggles to harmonize with the given image prior, we derive the optimal initialization from the forward diffusion process $x_t = \sqrt{\bar{\alpha}_t}x_0 + \sqrt{1-\bar{\alpha}_t}\epsilon$. For inpainting, the complete image $x_0$ consists of the observed region $y$ and the true masked region $x^m$. Therefore, the theoretically optimal starting point at timestep $T$ is:}
\begin{equation}
{x_T^* = \sqrt{\bar{\alpha}_T}(y + x^m) + \sqrt{1-\bar{\alpha}_T}\epsilon.}
\label{optimal xt}
\end{equation}

{Though $\bar{\alpha}_T$ is small, it remains strictly positive, ensuring that $x_T^*$ inherently preserves a residual signal from the true missing region $x^m$. Vanilla diffusion, however, initializes the masked region with pure standard Gaussian noise $\mathcal{N}(0, \mathbf{I})$. This practice mathematically eradicates the crucial signal term $\sqrt{\bar{\alpha}_T}x^m$, leading to deviated sampling trajectory. Thus, we believe that an initialization with a smaller distribution discrepancy is strictly needed for robust and coherent inpainting.}

To verify our hypothesis that an initialization with smaller distributional discrepancy lead to improved free-form inpainting, we conduct a series of analyses comparing different initialization strategies. These include: (1) vanilla Gaussian noise, (2) sampling from a Gaussian distribution fit on the unmasked region of given image, (3) Gaussian mixture fit on the unmasked region of given image, and (4) sampling from the real distribution of the masked region (which is ideal). 
We use the Kullback–Leibler (KL) divergence distribution plots to illustrate the difference between distributions on 100 images from ImageNet 1K. As shown in Fig.~\ref{motivation}, as the KL divergence gradually decreases——indicating the initialization transitions from incompatible to harmonious, the inpainting results evolve from unsatisfactory to high-quality. These results validate that initializations with smaller distributional discrepancies yield better inpainting outcomes.


{\textbf{Validity of the GMM Assumption.} We assume that the statistical distribution of the unmasked regions can serve as a robust proxy for that of the masked regions. This assumption is fundamentally grounded in two key observations. First, natural images inherently exhibit strong \textbf{spatial continuity and local coherence}. Adjacent regions typically reside on a shared feature manifold with consistent textures and colors, rendering the visible context a highly reliable prior for estimating the global image distribution. Second, the primary objective of free-form inpainting is to reconstruct missing areas harmoniously with their surroundings, rather than \textbf{hallucinating semantically divergent or statistically independent objects}. Therefore, constraining the initial generation seed to the unmasked distribution mathematically aligns with the requirement of visual consistency. Furthermore, this assumption does not restrict our method to unguided scenarios. As validated in Section IV-E, when external text prompts are introduced, our distribution-based initialization acts as a \textbf{structural anchor}. It synergizes effectively with the textual conditioning, consistently outperforming baselines by ensuring that the prompt-guided generation remains seamlessly grounded in the original image context.}
\label{headings}
\subsection{Primary Seed Initialization}
\label{primary}
Based on the insight proposed in Sec. \ref{moti}, we propose a tentative seed initialization that infers the masked image distribution, thereby encouraging the model to generate more consistent content. 
Specifically, we use the unmasked region to estimate the overall image distribution, which might closely match the distribution of the masked area. 
We employ a Gaussian Mixture Model (GMM) to fit the distribution of the unmasked image as:
\begin{equation}
    p(x_i|\theta_k)=\sum_{k=1}^K\pi_kN(x_i|\mu_k,\Sigma_k), i=1,\cdots,N \label{4}
\end{equation}
where $x$ denotes a pixel in the image $X$, which contains RGB values $r$, $g$, and $b$. 
Based on the above assumptions, if we can derive approximately close enough to $\mu$ and $\Sigma$ in Eq.~(\ref{4}) from the area outside the mask, we can sample an ideal initialization seed from the distribution. Therefore, we define the approximate distribution as:
\begin{equation}
        x_{\texttt{masked}} \sim \sum_{k=1}^K\hat{\pi}_kN(x_i|\hat\mu_k,\hat\Sigma_k), \label{6}
\end{equation}
where $\hat{\pi}_k$, $\hat{\mu}_k$ and $\hat{\Sigma}_k$ represent the mixing proportion, mean and variance of the $k$-th component in the GMM, respectively.

Then, we concatenate the random seed $x_{\texttt{masked}}$ sampled from the approximate distribution with the areas outside the mask $x_{\texttt{unmasked}}$ to obtain the primary seed initialization $x_{ini}$ by $x_{ini}=x_{\texttt{masked}}\cdot m + x_{\texttt{unmasked}}\cdot (1-m)$, where $m$ is a binary metrics used to represent the inpainting mask. Then we combine the seed with a slight level of random noise. Due to the additional randomness incorporated into the generation process, greater diversity is guaranteed. We discuss the diversity of IS-Diff in Sec.~\ref{4.4}. As illustrated in Fig.~\ref{method} (a), the primary initialized seed processes a data distribution that is more similar to that of a human face. With a statistically closer distribution, the model is more likely to harmonize the masked and unmasked regions, thus generating results that resemble the original regions.


\renewcommand{\algorithmicrequire}{\textbf{Input:}}

\begin{algorithm}[!t]
    \caption{Inpainting using IS-Diff }
    \label{Algorithm}
    \begin{algorithmic}
        \REQUIRE  checkpoint timestep for dynamic selective refinement $t_c$, maximum number of iterations $n$, dynamic adjustment strength $\Delta t$, DCE threshold $\epsilon$, sampling steps $T$, input image $x$, input mask $m$\\
        \STATE \textcolor{gray}{// Primary Seed Initialization}
        \STATE $x_{\texttt{masked}}\sim GMM(\hat{\pi},\hat{\mu}, \hat{\Sigma})$ (Eq.~\ref{6})
        \STATE $x_{ini}=x_{\texttt{masked}}\cdot m+x_{\texttt{unmasked}}\cdot(1-m)$ 
        \STATE $\hat{t} = T, k=0$
        \STATE {$x_{\hat{t}}=\sqrt{\overline{\alpha}_{\hat{t}}}x_{ini}+\sqrt{1-\overline{\alpha}_{\hat{t}}\varepsilon}$ (Eq.~\ref{8})}
        \WHILE{$k<n$}
        \FOR{$t=T$ to $1$}
        \STATE Sample $x_{0|t},x_{t-1}$ using baseline model 
        \STATE \textcolor{gray}{// dynamic selective refinement}
        \STATE Calculate DCE score using $x_{0|\hat{t}}$ when $t=t_c$
        \IF{$DCE>\epsilon$}
        %
        \STATE $k = k+1,\hat{t}=\hat{t}-\Delta t$
        \IF{$k=n$}
        \STATE $x_{0|t_c}=\arg\min\,DCE(x^k_{0|t_c})$
        \ELSE
        \STATE \textbf{break}
        \ENDIF
        
        
        \ENDIF
        
        \ENDFOR
        \ENDWHILE
        
        \RETURN $x_0$
    \end{algorithmic}
\end{algorithm}


\subsection{Dynamic Selective Refinement}
\label{3.3}
Although prior works have explored ways to encourage harmonious content from unmasked regions, the stochastic nature of diffusion models often causes error accumulation during inpainting. Recent training-free approaches s~\cite{Wang_Yu_Zhang_2022, Lugmayr_Danelljan_Romero_Yu_Timofte_Van_Gool_2022} attempt to mitigate this by reproducing the diffusion process, but this incurs prohibitive computational costs. Moreover, when internal conflicts are severe, repeated generation can even amplify such conflicts, frequently resulting in inpainting failures (e.g., third column, second row in Fig. 2).
Consequently, instead of refining inpainting with previous results which could mislead the model, we propose a dynamic selective refinement mechanism to select a better initialization. In essence, our approach can be regarded as a selection-based form of noise optimization. However, recent selection-based methods are designed for T2I task and rely on specific text prompt, which make them unsuitable for free-form inpainting. Furthermore, selection-based methods often need gradient computation in every inference step, leading to extra computational cost. Specifically, this mechanism iteratively evaluates intermediate inpainting results of an appropriate ``checkpoint" timestep to determine whether the masked and unmasked regions are distributionally harmonious. If not, we increase the strength of initialization to noise to refine the results accordingly. 

To assess the harmonization quality of inpainting results, we employ the Cross-Entropy between the histograms of intensity/texture Distributions (DCE) in the masked and unmasked regions as a metric:
\begin{equation}
    {DCE = -\sum_{i} H_{\text{masked}}(i) \log H_{\text{unmasked}}(i),}
\label{eq:cross_entropy_metric}
\end{equation}
{where $H_{\text{masked}}$ and $H_{\text{unmasked}}$ denote the histogram distributions of pixel values in masked and unmasked regions, respectively, and $i$ indexes the histogram bins (we use 64 bins per RGB channel).}
 In the inpainting procedure, if the above-mentioned DCE metric is above a pre-defined threshold $\epsilon$, we consider the inpainting to be unsatisfactory and regenerate the inpainting. When the maximum number of attempts is reached, we adopt the latent variable $x_t$ from the iteration with the minimal DCE value and continue the generation process.

Furthermore, we adjust the strength of our proposed primary seed initialization dynamically on the basis of DCE metrics. We reverse primary see d to the timestep $\hat{t}$ by,
\begin{equation}
    x_{\hat{t}}=\sqrt{\overline{\alpha_{\hat{t}}}}{x_{ini}}+\sqrt{1-\overline{\alpha_{\hat{t}}}}{\varepsilon},\label{8}
\end{equation}
{where $x_{ini}$ denotes the primary seed initialization defined in Sec.~\ref{primary}, and} $\hat{t}=T$ {initially}. Every time the inpainting fails to harmonize the distributional discrepency, we adjust the noise adding timestep to $\hat{t} = \hat{t} - \Delta t$, where $\Delta t$ is a hyperparameter that controls the strength of adjustment. By instituting a ``checkpoint'' mechanism for the inpainting process, our method facilitates the detection of generation deviations from the unmasked content while concurrently achieving significant reduction in the considerable time budget in prior methods.

\section{Experiment}
\label{sec:Experiment}
\subsection{Experiments Setup}
\label{4.1}

\textbf{Benchmark.} We validate our method on the CelebA-HQ 1K~\cite{Liu_Luo_Wang_Tang_2015} dataset and ImageNet 1K~\cite{Russakovsky_Deng_Su_Krause_Satheesh_Ma_Huang_Karpathy_Khosla_Bernstein_et_al._2015} dataset. We use the 1000 validation images from CelebA-HQ and ImageNet separately. {We also utilize 5000 validation images from Places2~\cite{zhou2017places} for additional evaluation on high resolution images in Sec.~\ref{high reso}.} For masks, we use LaMa \cite{Suvorov_Logacheva_Mashikhin_Remizova_Ashukha_Silvestrov_Kong_Goka_Park_Lempitsky_2022} setting for ``Wide'' masks and adopt the same ``Half'' and ``Expand'' mask settings as in RePaint
\cite{Lugmayr_Danelljan_Romero_Yu_Timofte_Van_Gool_2022} to verify the performance of our method under extreme large mask conditions. ``Half'' masks provide the left half of the image as input and ``Expand'' masks leave only the central 64x64 portion of a 256$\times$256 image.

  \begin{figure*}[tbh]
    \centering
    
    \includegraphics[width=0.9\linewidth]{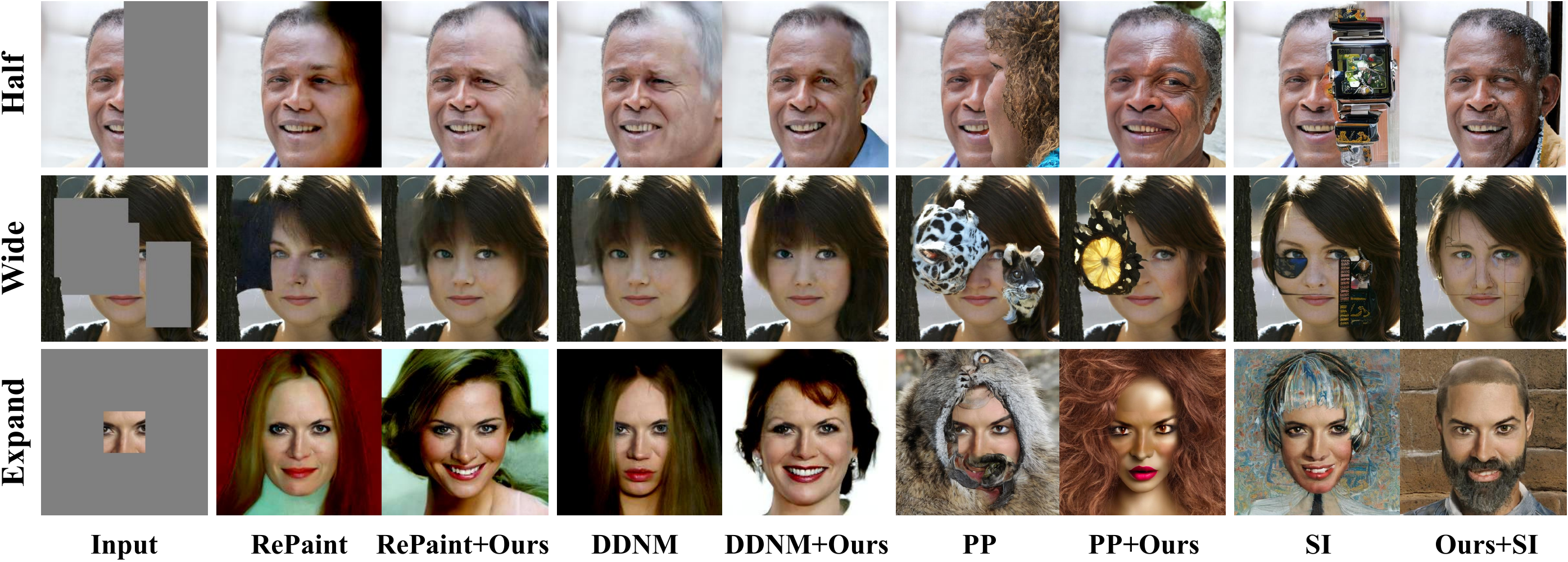}
    \caption{Qualitative comparison with state-of-the-art methods on CelebA-HQ. ``SI'' denotes Stable Diffusion In painting, and ``PP'' denotes PowerPaint.}
    \label{CelebA}
  \end{figure*}

  \begin{figure*}[tbh]
    \centering
    
    \includegraphics[width=0.9\linewidth]{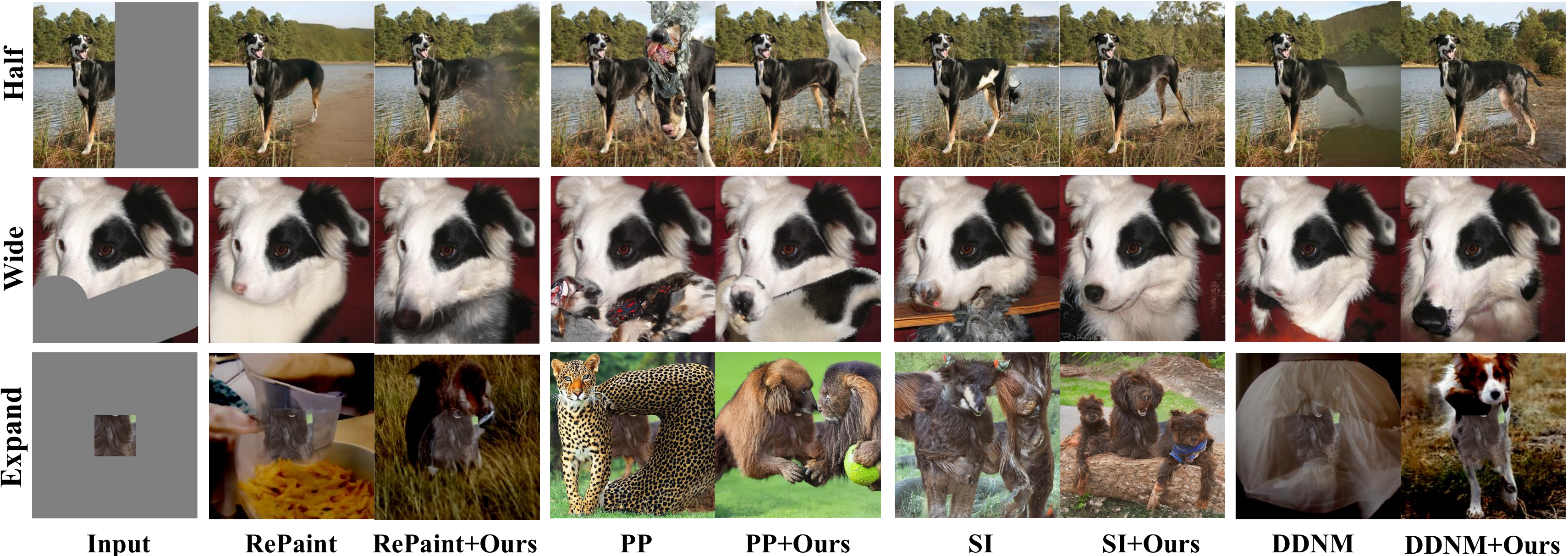}
    \caption{Qualitative comparison with state-of-the-art methods on ImageNet. ``SI'' denotes {Stable Diffusion Inpainting}, ``PP'' denotes PowerPaint.}
    \label{ImageNet}
  \end{figure*}
  


\noindent \textbf{Compared Methods.} To evaluate our method, we compare IS-Diff with current state-of-the-art training-free diffusion-based methods for image inpainting, including DDNM \cite{Wang_Yu_Zhang_2022}, RePaint \cite{Lugmayr_Danelljan_Romero_Yu_Timofte_Van_Gool_2022} and methods based on latent diffusion models including {Stable Diffusion Inpainting} 2.0~\cite{Rombach_Blattmann_Lorenz_Esser_Ommer_2022}, PowerPaint~\cite{zhuang2023task}, and HD Painter~\cite{manukyan2023hd}. Among them, {Stable Diffusion Inpainting} is a pre-trained baseline, DDNM, Repaint and HD-Painter are training-free methods, while PowerPaint requires finetuning.  {We also evaluate IS-Diff on modern flow matching baselines on FLUX-Fill and FLUX-Inpaint~\cite{labs2025flux1kontextflowmatching}.}

\begin{table*}[thb]
\small
\begin{center}
\caption{Comparison with state-of-the-art methods on ImageNet 1K. The LPIPS and FID scores are reported. Lower is better. \textbf{Bold} fonts represent the best result. ``HP'' denotes HD Painter while ``PP'' denotes PowerPaint.}
\label{tab:state-of-the-art-imagenet}
\begin{tabular}{cclclclc}
\hline
\multirow{2}{*}{Methods}    & \multicolumn{2}{c}{Wide}                               & \multicolumn{2}{c}{Half}                               & \multicolumn{2}{c}{Expand}                              & \multirow{2}{*}{\begin{tabular}[c]{@{}c@{}}Time Cost\\    (s/image)↓\end{tabular}} \\ \cline{2-7}
                            & \multicolumn{1}{l}{LPIPS↓} & FID↓                      & \multicolumn{1}{l}{LPIPS↓} & FID↓                      & \multicolumn{1}{l}{LPIPS↓} & FID↓                       &                                                                                    \\ \hline
SI~\textbf{[CVPR 22]}~\cite{Rombach_Blattmann_Lorenz_Esser_Ommer_2022}      & 0.273                      & 70.44                     & 0.385                      & 66.66                     & 0.763                      & 139.76                     & 5                                                                                  \\
SI + HP~\textbf{[ICLR 25]}~\cite{manukyan2023hd} & 0.249                      & \multicolumn{1}{c}{64.82} & 0.463                      & \multicolumn{1}{c}{92.17} & 0.781                      & \multicolumn{1}{c}{170.18} & 15                                                                                 \\ 
\textbf{SI + Ours}          & \textbf{0.238}             & \textbf{54.91}            & \textbf{0.381}             & \textbf{61.32}            & \textbf{0.676}             & \textbf{123.36}            & \textbf{3}                                                                         \\ \hline
PP~\textbf{[ECCV 24]}~\cite{zhuang2023task} & 0.263                      & \multicolumn{1}{c}{63.83} & 0.397                      & \multicolumn{1}{c}{91.62} & 0.779                      & \multicolumn{1}{c}{178.45} & \textbf{3}                                                                         \\
\textbf{PP + Ours} & \textbf{0.254}                      & \multicolumn{1}{c}{\textbf{53.85}} & \textbf{0.373}                      & \multicolumn{1}{c}{\textbf{68.69}} & \textbf{0.707}                      & \multicolumn{1}{c}{\textbf{160.62}} & \textbf{3}                                                                         \\ \hline
DDNM'~\textbf{[ICLR 23]}~\cite{Wang_Yu_Zhang_2022}    & 0.130                      & 23.64                     & 0.358                      & 42.09                     & 0.790                      & 145.08                     & 21                                                                                 \\
\textbf{DDNM + Ours}        & \textbf{0.128}             & \textbf{22.01}            & \textbf{0.328}             & \textbf{39.94}            & \textbf{0.721}             & \textbf{121.17}            & \textbf{4}                                                                         \\ \hline
RePaint'~\textbf{[CVPR 22]}~\cite{Lugmayr_Danelljan_Romero_Yu_Timofte_Van_Gool_2022} & 0.134                      & \multicolumn{1}{c}{25.01} & 0.351                      & \multicolumn{1}{c}{42.95} & 0.777                      & \multicolumn{1}{c}{143.00} & 22                                                                                 \\
\textbf{RePaint + Ours}    & \textbf{0.131}                      & \textbf{23.34}                    & \textbf{0.332}                      & \textbf{40.47 }                   & \textbf{0.733}                      & \textbf{122.78}                                                                        & \textbf{4}                                                                         \\ \hline
{{} FLUX.1 Fill~\cite{labs2025flux1kontextflowmatching}} & {{} 0.117} & {{} 29.75} & {{} 0.316} & {{} 38.16} & {{} 0.650} & {{} 108.49} & {{} 10} \\
{{} \textbf{FLUX.1 Fill + Ours}} & {{} \textbf{0.113}} & {{} \textbf{28.31}} & {{} \textbf{0.304}} & {{} \textbf{36.42}} & {{} \textbf{0.641}} & {{} \textbf{93.68}} & {{} \textbf{8}} \\ \hline

{FLUX-Inpaint~\cite{labs2025flux1kontextflowmatching}} & {0.130} & {26.11} & {0.309} & {62.99} & {0.652} & {145.95} & {\textbf{10}} \\
{\textbf{FLUX-Inpaint + Ours}} & {\textbf{0.121}} & {\textbf{23.07}} & {\textbf{0.287}} & {\textbf{47.87}} & {\textbf{0.642}} & {\textbf{119.08}} & {\textbf{10}} \\

\hline
\end{tabular}
\end{center}
\end{table*}

\noindent \textbf{Quantitative Metrics.} Since L1 and L2 distances are not suitable for evaluating the effectiveness of inpainting since multiple natural completions are possible, we use the commonly reported perceptual metrics LPIPS and FID as the main quantitative metrics to compare IS-Diff with baseline methods.\\
\textbf{Implementation Details.} For CelebA-HQ 1K, we use the 256$\times$256 pre-trained model trained on CelebA-HQ by \cite{meng2022sdedit}. For ImageNet 1K, we utilize the 256$\times$256 pre-trained model provided by \cite{Dhariwal_Nichol_2021}. To ensure a fair comparison, we use the same pre-trained model for all compared diffusion methods. We employ DDIM~\cite{Song_Meng_Ermon_2020} as the base sampling strategy with 100 steps for all diffusion methods. For ``Wide'' and ``Half'' masks, the repeat time and the jump size of the resampling strategy are set to $10$ for DDNM and RePaint. For ``Expand'' masks, we set the parameters to $l=4$ and $j=4$. We set the maximum number of iterations for {Dynamic Selective Refinement} to $3$, $t_c=0.6T$. The threshold $\epsilon$ is set to $2.5$ and the components $K$ of GMM are set to $5$ for all settings. Furthermore, we implement IS-Diff into Stable Diffusion to better discuss the effectiveness of our method on latent features using {Stable Diffusion Inpainting} 2 pre-trained model with 50 sampling steps. Specifically, we apply our primary seed initialization in latent space, while the dynamic selective refinement is implemented to the RGB space. Other methods based on stable diffusion use 100 sampling steps instead to ensure the evaluation under similar computational cost. All experiments are executed on a single NVIDIA 4090 GPU for once.
\subsection{Comparison with State-of-the-Art methods}
\label{4.2}
\begin{table*}[thb]
\begin{center}
\small
\caption{Comparison with state-of-the-art methods on CelebA-HQ 1K. The LPIPS and FID scores are reported. Lower is better. \textbf{Bold} fonts represent the best result. ``HP'' denotes HD Painter while ``PP'' denotes PowerPaint. As reported by Lugmayr \textit{et al.}~\cite{Lugmayr_Danelljan_Romero_Yu_Timofte_Van_Gool_2022}, LPIPS is not an suitable metric for large masks because diverse solutions different from ground truth are possible. }
\label{tab:state-of-the-art-celeba}
\begin{tabular}{cccccccc}
\hline
\multirow{2}{*}{Methods}    & \multicolumn{2}{c}{Wide}                              & \multicolumn{2}{c}{Half}                              & \multicolumn{2}{c}{Expand}                            & \multirow{2}{*}{\begin{tabular}[c]{@{}c@{}}Time Cost\\    (s/image)↓\end{tabular}} \\ \cline{2-7}
                            & \multicolumn{1}{l}{LPIPS↓} & \multicolumn{1}{l}{FID↓} & \multicolumn{1}{l}{LPIPS↓} & \multicolumn{1}{l}{FID↓} & \multicolumn{1}{l}{LPIPS↓} & \multicolumn{1}{l}{FID↓} &                                                                                    \\ \hline
SI~\textbf{[CVPR 22]}~\cite{Rombach_Blattmann_Lorenz_Esser_Ommer_2022}      & 0.237                      & 107.07                   & 0.323                      & 91.08                    & 0.617                      & 165.18                   & 5                                                                                  \\
SI + HP~\textbf{[ICLR 25]}~\cite{manukyan2023hd} & 0.207                      & 66.30                    & 0.458                      & 164.69                   & 0.686                      & 230.50                   & 15                                                                                 \\
\textbf{SI + Ours}         & \textbf{0.191}             & \textbf{63.22}           & \textbf{0.303}             & \textbf{52.88}           & \textbf{0.609}             & \textbf{132.64}          & \textbf{3}                                                                         \\ \hline
PP~\textbf{[ECCV 24]}~\cite{zhuang2023task}      & 0.223                      & 101.90                   & 0.267                      & 53.32                    & 0.691                      & 276.22                   & \textbf{3}                                                                         \\
\textbf{PP + Ours}         & \textbf{0.207}             & \textbf{74.96}           & \textbf{0.254}             & \textbf{42.84}           & \textbf{0.667}             & \textbf{262.46}          & \textbf{3}                                                                         \\ \hline
DDNM~\textbf{[ICLR 23]}~\cite{Wang_Yu_Zhang_2022}    & 0.081                      & 14.12                    & 0.236                      & 23.07                    & 0.567                      & 94.23                    & 21                                                                                 \\
\textbf{DDNM + Ours}       & \textbf{0.081}             & \textbf{13.80}           & \textbf{0.191}             & \textbf{23.02}           & \textbf{0.547}             & \textbf{60.86}           & \textbf{4}                                                                         \\ \hline
RePaint~\textbf{[CVPR 22]}~\cite{Lugmayr_Danelljan_Romero_Yu_Timofte_Van_Gool_2022} & 0.083                      & 14.42                    & 0.248                      & 24.47                    & 0.564                      & 99.21                    & 22                                                                                 \\
\textbf{RePaint + Ours}    & \textbf{0.079}             & \textbf{13.82}           & \textbf{0.203}             & \textbf{23.79}           & \textbf{0.549}             & \textbf{63.39}           & \textbf{4}                                                                         \\ \hline
{{} FLUX.1 Fill~\cite{labs2025flux1kontextflowmatching}} & {{} 0.077}      & {{} 14.53}      & {{} 0.183}      & {{} 28.38}      & {{} 0.462}      & {{} 69.98}      & {{} 10}                                                                   \\
{{} \textbf{FLUX.1 Fill + Ours}} & {{} \textbf{0.073}} & {{} \textbf{13.00}} & {{} \textbf{0.172}} & {{} \textbf{26.39}} & {{} \textbf{0.451}} & {{} \textbf{55.93}} & {{} \textbf{8}}                                                       \\ \hline
\end{tabular}
\end{center}
\end{table*}

\begin{table}[thb]
\begin{center}

\scriptsize
\caption{User Study Results. We collect 600 responses from 30 participants to compare the visual effects of IS-Diff against state-of-the-art diffusion-based methods on CelebA and ImageNet. The table shows that our method receives the majority of votes. \textbf{Bold} fonts represent the best result.}
\begin{tabular}{c|ccc|ccc}
\hline
      \multirow{2}{*}{Methods} & \multicolumn{3}{c|}{CelebA-HQ}     & \multicolumn{3}{c}{ImageNet}       \\ 
& Wide& Half& Expand& Wide& Half& Expand\\ \hline
DDNM~\cite{Wang_Yu_Zhang_2022}& 34.4\%& 16.9\%& 24.9\%& 21.3\%& 25\%& 22.6\%\\
RePaint~\cite{Lugmayr_Danelljan_Romero_Yu_Timofte_Van_Gool_2022}& 16.8\%& 23.5\%& 36.3\%& 21.8\%& 19.9\%& 25.1\%\\
\textbf{Ours}& \textbf{48.8\%}& \textbf{59.6\%}& \textbf{38.8\%}& \textbf{56.9\%}& \textbf{55.1}\%& \textbf{52.3}\%\\ \hline
\end{tabular}
\label{tab: User Study}
\end{center}
\end{table}
\textbf{Quantitative Comparison.} 
We validate the effectiveness of our method through experiments on ImageNet 1K and CelebA-HQ 1K, with results summarized in Table~\ref{tab:state-of-the-art-imagenet} and Table~\ref{tab:state-of-the-art-celeba}. 
To highlight the effectiveness of our method as a plug-and-play module, we integrate our method into multiple baseline methods and compare with relative methods respectively. Specifically, we use {Stable Diffusion Inpainting} and DDNM as baselines respectively.

For {Stable Diffusion Inpainting}, we compare our IS-Diff with methods based on {Stable Diffusion Inpainting} pre-trained model, including HD Painter and PowerPaint. On the other hand, we compare our method with DDNM and RePaint for DDNM baseline because they have been proven to be equivalent in most cases~\cite{Wang_Yu_Zhang_2022}.

Results demonstrate that IS-Diff consistently enhances performance across various baseline frameworks, validating its plug-and-play effectiveness. Within the DDNM-based comparison group, IS-Diff outperforms RePaint and vanilla DDNM across all metrics and mask configurations when using same pre-trained models and even less sampling steps—this consistent superiority underscores its ability to boost performance within the DDNM framework.

In the {Stable Diffusion Inpainting}-based comparison group, our method also outperforms existing approaches built on the same pre-trained model, especially in the challenging ``Expand'' setting, highlighting its effectiveness as a plug-and-play module. Notably, {Stable Diffusion Inpainting}-based methods exhibit unique behavioral patterns: across both datasets, they achieve stronger FID scores in the "Half" mask setting compared to "Wide", likely due to the greater generative flexibility afforded by "Half" configurations, which aligns well with the strengths of {Stable Diffusion Inpainting} models. Although PowerPaint~\cite{zhuang2023task}  outperforms our method in terms of LPIPS in ``Half " setting, but we have to report that PowerPaint requires fine-tuning on {Stable Diffusion Inpainting} model, which is different from other training-free methods.

To further validate the generalizability and plug-and-play nature of IS-Diff, we conduct additional experiments using the current state-of-the-art inpainting model, FLUX.1 Fill~\cite{labs2025flux1kontextflowmatching}. As demonstrated in Table~\ref{tab:state-of-the-art-imagenet} and~\ref{tab:state-of-the-art-celeba}, the integration of IS-Diff consistently yields performance gains across all masking scenarios and datasets, including ImageNet-1K and CelebA-1K. Notably, our method significantly refines the generative quality in the most challenging ``Expand'' settings, proving that the proposed distribution-level guidance effectively compensates for limited contextual information even in highly optimized modern diffusion architectures.

These results confirm that IS-Diff serves as a robust, architecture-agnostic enhancement for high-fidelity image completion.
Together, these results confirm that IS-Diff can enhance the inpainting performance of baselines with different architectures, validating its universal value as a plug-and-play module.

\textbf{Qualitative Comparison.} We report the qualitative results on ImageNet and CelebA-HQ in Fig.~\ref{CelebA} and Fig.~\ref{ImageNet}, respectively. We observe that on ImageNet, our IS-Diff demonstrates significantly better performance, especially when facing hard cases. On CelebA-HQ, RePaint, DDNM, and our method all achieve satisfying results with the frontal face image. However, when the image shows the side face (first row in Fig.~\ref{CelebA}), only our method produces a rational inpainting. This is because the diffusion model is pre-trained on mainly frontal face images, and the previous methods use random initial seeds in the mask region so it is hard to achieve consistent results. Our approach also shows improved performance with extensively masked images (see the third row in Fig.~\ref{CelebA}) because of the reduced conflict between the unmasked region and the remaining heavily masked section.\\
\textbf{Human Study.} We conduct a user study to compare the inpainting quality of our method against the state-of-the-art diffusion-based methods. Users were shown the masked images along with the generated results from different methods and were asked to select the image they considered the most realistic. To avoid bias in the user study, the methods were presented anonymously, and the order of image presentation was randomized. The user study evaluated 20 sets of images randomly selected from each dataset and mask setting. We collected responses from 30 participants, resulting in 600 votes. Table \ref{tab: User Study} shows that our method receives the majority of votes, reflecting that our method can achieve more harmonious and realistic results.
\begin{figure*}[t]
     \centering
    
     \includegraphics[width=0.95\linewidth]{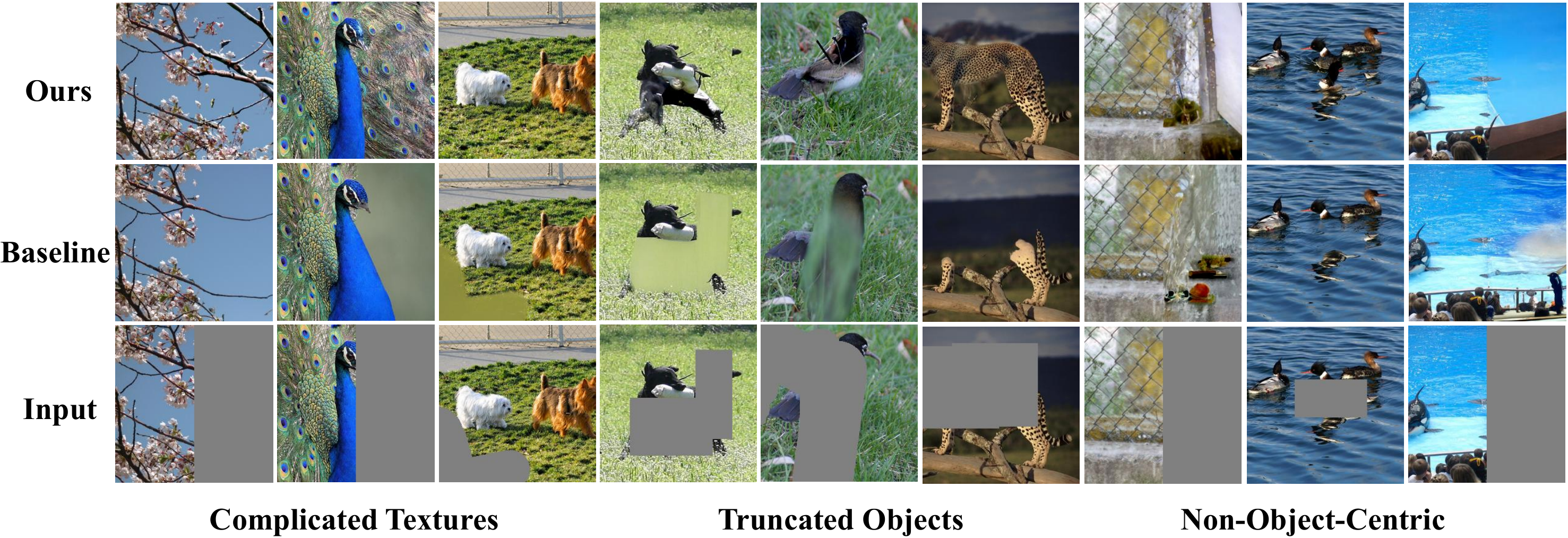}
      \caption{Results on challenging scenarios including complicated textures, non-object-centric, and object-truncated scenes. Our method showcasing better inpainting quality and more reasonable content.}
     \label{fig:challenging-scene}
 \end{figure*}
\subsection{Evaluation on high resolution dataset}
\label{high reso}
\begin{figure}
    \centering
    \includegraphics[width=\linewidth]{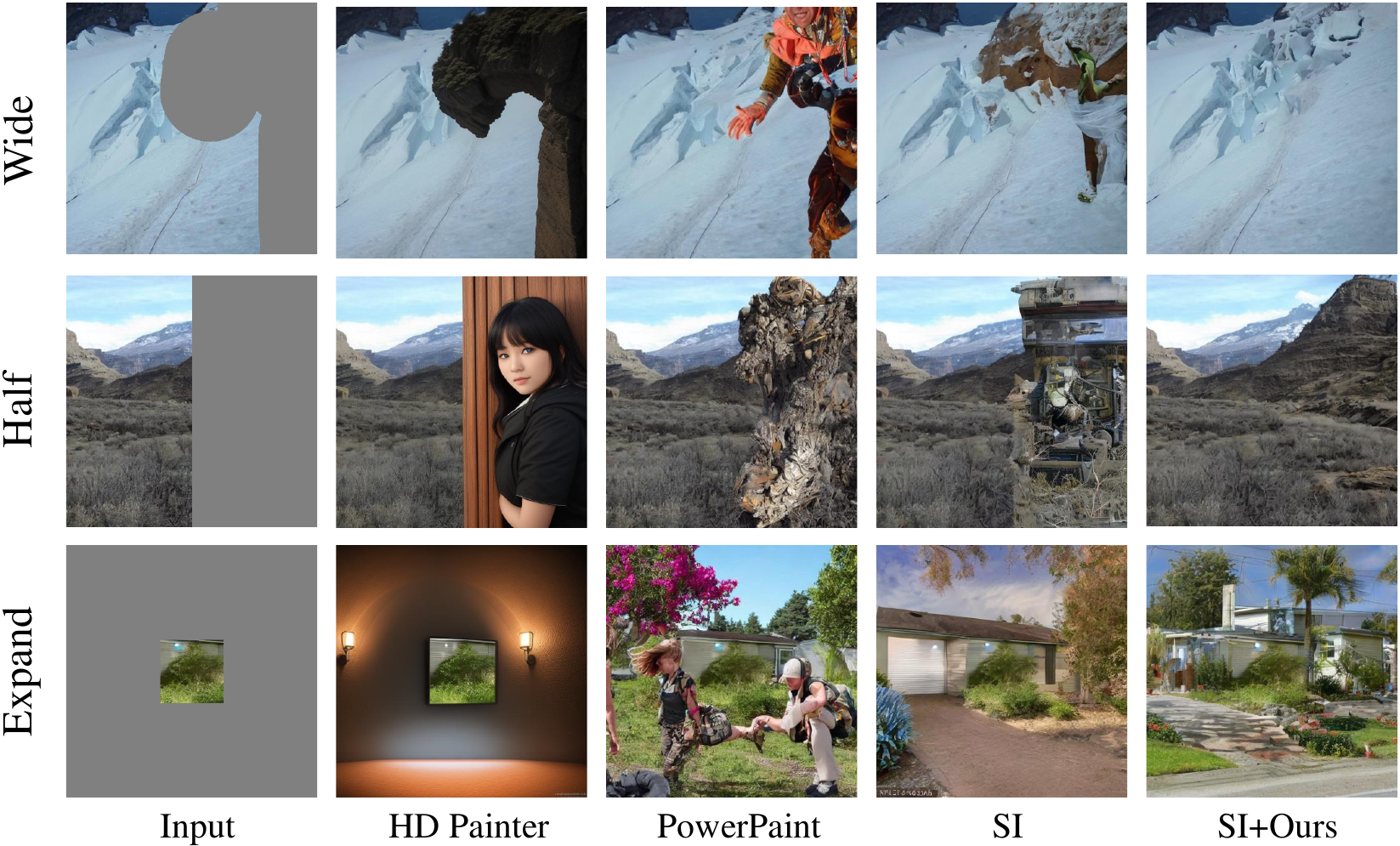}
    \caption{Qualitative comparison with state-of-the-art methods on Places2. ``SI'' denotes {Stable Diffusion Inpainting}.}
    \label{fig:places2}
\end{figure}

To further validate the scalability of our method, we evaluate our method on the Places2 dataset with 512×512 resolution. We utilize 5000 images from the validation set of Places2 and compare our methods based on {Stable Diffusion Inpainting} with other stable diffusion based methods. Both quantitative and qualitative results demonstrate the effectiveness of our method. As shown in Table~\ref{high-res}, our proposed IS-Diff outperforms current state-of-the-art methods in all settings and metrics. We also report the qualitative results on Places2 in Fig.~\ref{fig:places2}. We observe that our method generates inpainting results with reasonable semantic content and coherent details compared to state-of-the-art methods. 
\begin{table*}[tb]
\small
\centering
\caption{Comparison with state-of-the-art methods on high resolution dataset. We use $512 \times 512$ images from Places2 dataset. The LPIPS and FID scores are reported. Lower is better. \textbf{Bold} fonts represent the best result. ``HP'' denotes HD Painter while ``PP'' denotes PowerPaint. }
\begin{tabular}{cccccccc}
\hline
\multirow{2}{*}{Methods}    & \multicolumn{2}{c}{Wide}       & \multicolumn{2}{c}{Half}        & \multicolumn{2}{c}{Expand}      & \multirow{2}{*}{\begin{tabular}[c]{@{}c@{}}Time Cost\\    (s/image)↓\end{tabular}} \\ \cline{2-7}
                            & LPIPS↓         & FID↓          & LPIPS↓         & FID↓           & LPIPS↓         & FID↓           &                                                                                    \\ \hline
SI'~\textbf{[CVPR 22]}~\cite{Rombach_Blattmann_Lorenz_Esser_Ommer_2022}      & 0.212          & 8.23          & 0.363          & 13.81          & 0.689          & 45.13          & 5                                                                                  \\
SI + HP'~\textbf{[ICLR 25]}~\cite{manukyan2023hd} & 0.234          & 12.43         & 0.469          & 74.57          & 0.779          & 105.46         & 15                                                                                 \\
SI + PP'~\textbf{[ECCV 24]}~\cite{zhuang2023task} & 0.267          & 25.09         & 0.423          & 64.14          & 0.711          & 87.64          & 3                                                                                  \\
\textbf{SI + Ours}          & \textbf{0.181} & \textbf{6.82} & \textbf{0.350} & \textbf{12.01} & \textbf{0.678} & \textbf{31.31} & \textbf{3}                                                                         \\ \hline
\end{tabular}

\label{high-res}
\end{table*}
\subsection{Complex Inpainting}
\label{4.3}
Our method improves inpainting performance in challenging cases, including complicated textures, non-object-centric and object-truncated scenarios. The complicated textures scenario poses a challenge for the model to generate masked regions with consistent textures, while the truncated‐object and non‐object‐centric scenarios require the model to obey stricter contextual constraints and produce plausible content. As shown in Fig.~\ref{fig:challenging-scene}, our method demonstrates stronger consistency with the original image’s texture and semantics. For example, our model generates more realistic peacock feathers and is capable of inpainting objects that have been truncated in the middle.
\subsection{Text-guided Inpainting}
To better validate the practicality of the proposed solution including text-guided inpainting, we conduct text-guided inpainting experiments on {EditBench~\cite{Wang_Saharia_Montgomery_Pont_Tuset_Noy_Pellegrini_Onoe_Laszlo_Fleet_Soricut_et_al._2022}}. Specifically, we utilize {Stable Diffusion Inpainting} as the baseline method and CLIP Score and SSIM metric as the quantitative metrics. Quantitaive results in Table~\ref{text-guided}, though not specially designed for text-guided inpainting, our method can improve the prompt alignment capability and generation quality of text-guided inpainting.
\begin{table}[t]
\centering
\caption{Quantitative comparison with {Stable Diffusion Inpainting} {on EditBench}.}
\label{text-guided}
\begin{tabular}{lcc}
\hline
\textbf{Method} & \textbf{CLIP Score $\uparrow$} & \textbf{SSIM $\uparrow$} \\
\hline
{Stable Diffusion Inpainting} & 30.1 & 0.708 \\
\textbf{{Stable Diffusion Inpainting} + Ours} & \textbf{30.7} & \textbf{0.726} \\
\hline
\end{tabular}

\end{table}
\subsection{{Qualitative results on flow-matching baseline}}

{We add qualitative visualization results of IS-Diff applied to the flow-matching-based model FLUX-Inpaint. The visual comparisons show that IS-Diff improves the contextual compatibility between the generated masked region and the visible image content, especially in challenging cases with large masked regions. These results further demonstrate that the proposed initialization strategy generalizes beyond conventional diffusion-based inpainting models.}

\begin{figure}[h]
    \centering
    \includegraphics[width=0.95\linewidth]{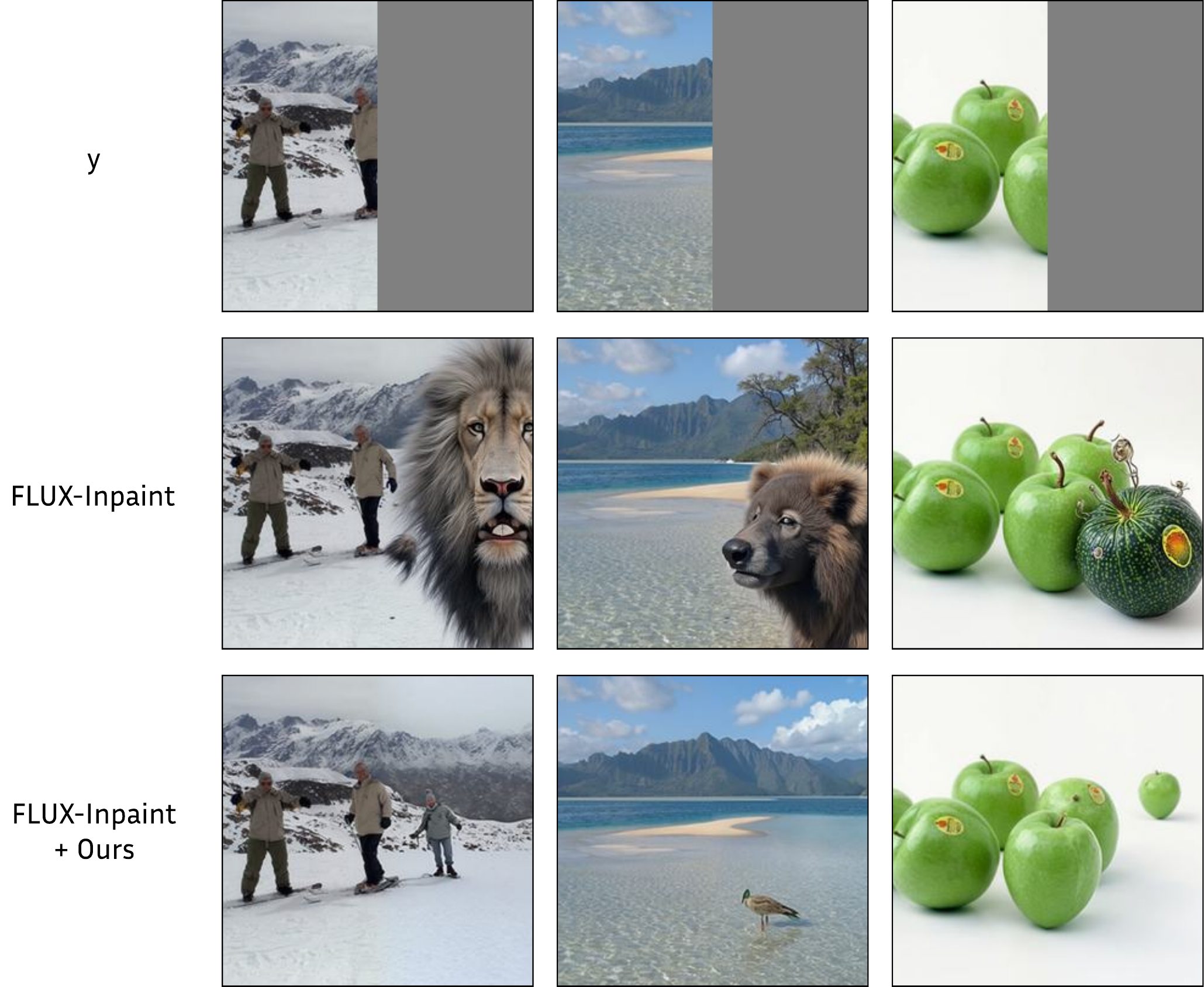}
    \caption{
    {Qualitative visualization results of IS-Diff applied to the flow-matching-based model FLUX-Inpaint. Compared with the baseline model, IS-Diff improves the contextual compatibility of the generated regions and reduces visually inconsistent artifacts.}
    }
    \label{fig:r4_flux_inpaint_visualization}
\end{figure}
\subsection{Ablation study}
\label{4.4}
\textbf{Primary Seed Initialization.} To evaluate the effect of the primary seed initialization, we perform ablation studies without the selective iterative refinement on the ImageNet dataset and compare the results with the DDNM baseline. As shown in Table \ref{tab:Ablation}, Ours (w/o DIR) consistently improves the baseline on all settings, and the improvement is especially significant on hard cases. These observations demonstrate that the primary initial seed successfully estimates the masked content, especially for the extremely large masks.

\begin{table*}[thb]
\begin{center}
\caption{{Quantitative ablation study on the ImageNet dataset. Here, ``DSR'' denotes the process of dynamic selective refinement. \textbf{Bold} fonts represent the best result. The quantitative results show that our method can achieve better performance even without DSR.}}

\small

\begin{tabular}{c|cc|cc|cc}
\hline
\multirow{2}{*}{Methods} & \multicolumn{2}{c|}{Wide} &  \multicolumn{2}{c|}{Half}& \multicolumn{2}{c}{Expand}\\
& LPIPS↓        & FID↓          & LPIPS↓        & FID↓          
& LPIPS↓        &FID↓          
\\ \hline
Baseline (w/o resample)& 0.160& 31.68& 0.351& 55.58& 0.780&127.69\\ 
{Ours (w/o DSR)}& 0.155& 30.02& 0.340& 50.47& 0.777&122.63\\
Ours& \textbf{0.128}& \textbf{22.01}& \textbf{0.328}& \textbf{39.94}& \textbf{0.721}& \textbf{121.17}\\
\hline
\end{tabular}

\label{tab:Ablation}
\end{center}
\end{table*}

\begin{figure}

     \centering
     \includegraphics[width=\linewidth]{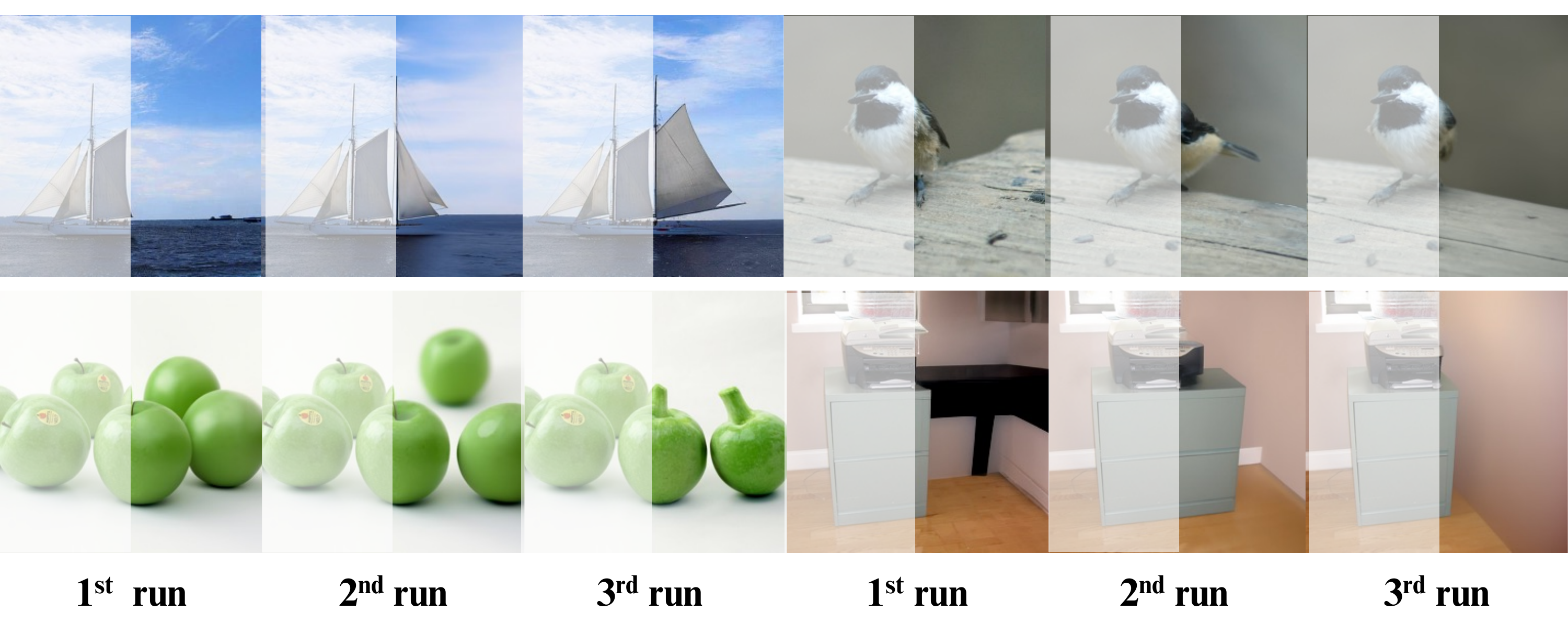}
     
     \caption{Diverse inpainting results on ImageNet with ``half'' masks. The right half of image is masked.}
     \label{fig:diversity}
\end{figure}
 \begin{table}[thb]
\small
\begin{center}
\caption{Quantitative comparison of L2 distance between results from multiple runs with the baseline model. Higher is better.}
\begin{tabular}{c|c}
\hline
Methods         & L2 Distance of two runs $\uparrow$ \\ \hline
Baseline (DDNM) & 0.2273              \\
Ours            & 0.3073              \\ \hline
\end{tabular}

\label{tab:diversity}
\end{center}
\end{table}

\begin{figure}
    \centering
    \includegraphics[width=\linewidth]{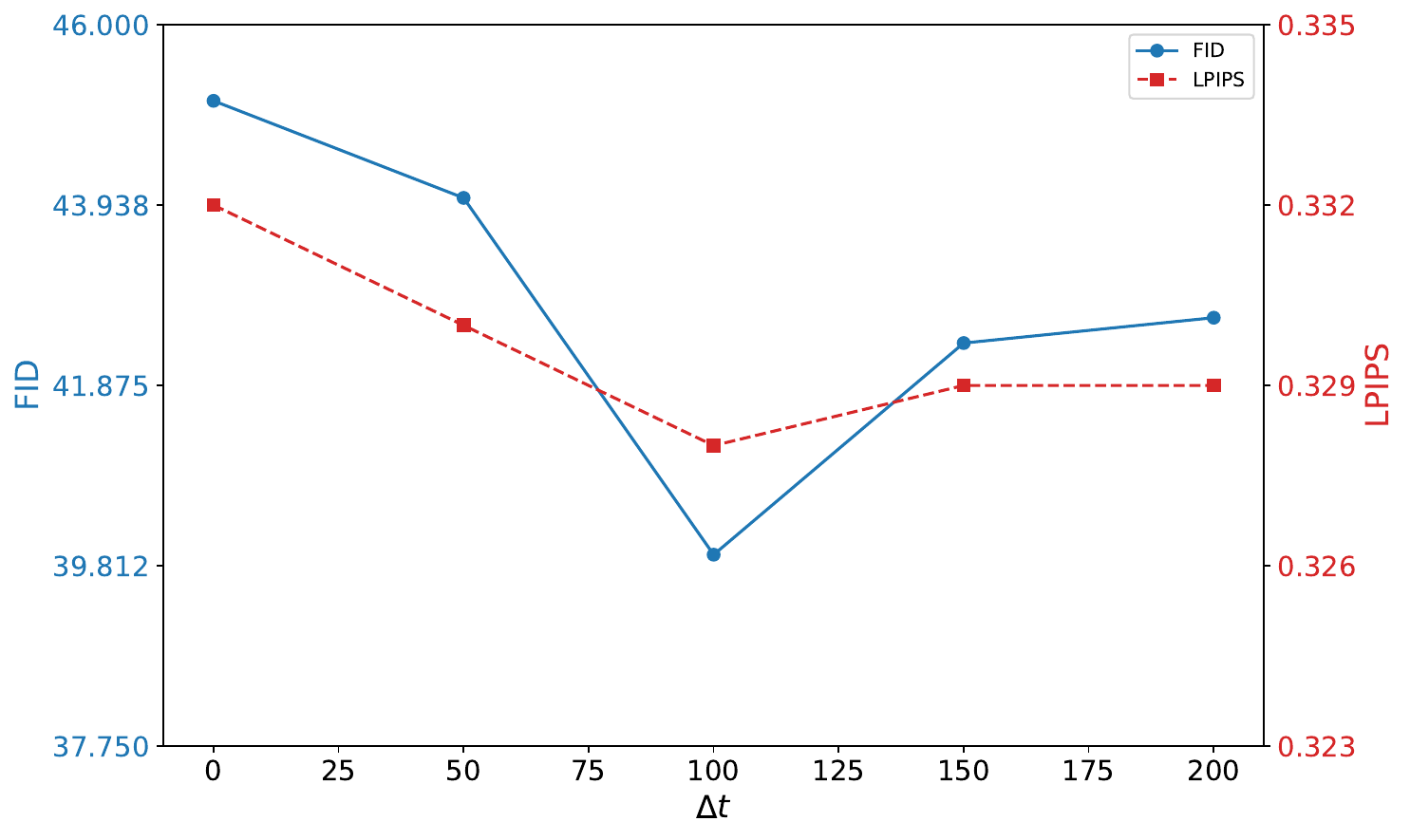}
    \caption{Comparison of different values of $\Delta t$.}
    \label{iterative refinement}
\end{figure}

\noindent{\textbf{Diverse Inpainting.} }
Diverse generation aims to generate reasonable results with different semantic information since multiple solutions are possible when the image is widely masked. 
We employ a dual assurance mechanism to ensure the diversity of our method. Our proposed primary seed initialization is not deterministic, but rather randomly sampled from the estimated distribution. Moreover, we integrate the initialization process with some noise to enhance the flexibility in generation. By preserving the randomness of initialization, our method can sample diverse results with enhanced inpainting quality (as shown in Fig.~\ref{fig:diversity}). We also conduct a quantitative comparison in Table \ref{tab:diversity}, showcasing our method is superior to baseline method in terms of diversity. {We use LPIPS distance between multiple runs to evaluate generation diversity.} Specifically, we sampled results twice from baseline and ours, and extract the AlexNet~\cite{NIPS2012_c399862d} features of those images. After that, we calculate the L2 feature distance between every two outputs of the same input. We regard the averaged L2 distance (which is equivalent to LPIPS) as an evidence of the diversity.
\\
\noindent{\textbf{Dynamic Adjustment.}} We conduct a supplementary ablation study in this section to analyze the impact of the strength of dynamic adjustment $\Delta t$,  we present ablation studies on ImageNet using half masks under five different settings in Fig.~\ref{iterative refinement}. We notice that the optimal quantitative outcomes occur when $\Delta t=100$.



\noindent \textbf{Comparison under Extended Time Consumption.} For extreme large mask conditions like ``expand'', it is challenging to generate satisfying results in a short time frame. Therefore, to better evaluate the robustness and efficiency of our method in extreme cases, we conduct additional experiments with the ``Expand'' mask under extended time conditions and compared our method with the state-of-the-art training-free method Repaint in terms of time consumption and quantitative metrics. The results are shown in Table \ref{time}. Specifically, for RePaint, we set $T=250$ and $r=15$ separately to provide more time for the model to generate harmonious results. For our method, we apply larger maximum iterations for dynamic selective refinement. 
We observe that, when given more time, both our method and Repaint achieve a better FID result. Our IS-Diff is observed to achieve better performance even with only 10\% the time consumption of RePaint ($T=250$). 
It is also worth noting that the LPIPS metric hardly changes with increased time consumption, which we believe is due to the lack of information under the ``Expand'' masks, causing the model to generate images that are semantically different from the original but still realistic, making the LPIPS metric not suitable for evaluation. This demonstrates that IS-Diff refines the inpainting effect in an efficient and flexible way. \\
\begin{table}[t]
\small
\caption{Experiment with more time conditions under the ``Expand'' setting on CelebA-HQ. 
Quantitative metrics and average time consumption are reported. Here, $n$ denotes the iteration number of iterative refinement, $T$ denotes the sampling steps and $r$ represents the repeat time of resampling in Repaint~\cite{Lugmayr_Danelljan_Romero_Yu_Timofte_Van_Gool_2022}. If not specified, we set $n$ to 1, $T$ to 100, and $r$ to 10 for  RePaint.}
\begin{center}

\begin{tabular}{c|ccc}
\hline
\multirow{2}{*}{Methods} & \multicolumn{2}{c}{Expand} &Inference time \\
               & LPIPS↓        & FID↓       &(s/image)↓\\ 
\hline
RePaint & 0.564& 99.21&21\\ 
RePaint ($T$=250)&               0.564&            86.22&50\\
RePaint ($r$=15)&               0.563&            102.51&27\\ \hline
 Ours ($n$=3) & 0.547& 60.86&4\\
 Ours ($n$=4)& 0.539& 57.82&5\\
 Ours ($n$=5)& 0.536& 57.09&5\\
\hline
\end{tabular}

\label{time}
\end{center}
\end{table}
{\textbf{Number of GMM Components ($K$).} We conduct an ablation study to analyze the impact of the number of GMM components $K$ on ImageNet with ``Half'' masks. As shown in Table~\ref{tab:k_ablation}, performance varies with different $K$ values. $K=5$ is used in our experiments as it achieves best performance.}

\begin{table}[h]
\small
\centering
\caption{{Ablation on number of GMM components $K$ on ImageNet (Half masks).}}
\label{tab:k_ablation}
\begin{tabular}{c|cc}
\hline
{$K$} & {LPIPS$\downarrow$} & {FID$\downarrow$} \\
\hline
{1} & {0.352} & {63.41} \\
{3} & {0.340} & {58.79} \\
{5} & {0.328}& {39.94}\\
{7} & {0.329}& {39.47}\\
 
\hline
\end{tabular}
\end{table}
\noindent{{\textbf{Checkpoint Timestep ($t_c$).} We analyze the sensitivity to the checkpoint timestep $t_c$, which determines when to evaluate the DCE metric during diffusion. An earlier checkpoint (smaller $t_c$) allows earlier detection of inconsistencies but may be less reliable due to higher noise levels. A later checkpoint (larger $t_c$) provides more stable estimates but leaves less time for refinement. Table~\ref{tab:tc} shows results on ImageNet (Half masks). We find $t_c = 0.6T$ achieves the best balance between detection reliability and refinement opportunity.}}

\begin{table}[h]
\small
\centering
\caption{{Sensitivity analysis of $t_c$ on ImageNet (Half masks).}}
\label{tab:tc}
\begin{tabular}{c|ccc}
\hline
{$t_c$} & {0.4T} & {0.6T} & {0.8T} \\
\hline
{LPIPS$\downarrow$} & {0.342} & {0.328} & {0.343}\\
{FID$\downarrow$} & {43.89} & {39.94} & {45.43}\\

\hline
\end{tabular}
\end{table}

\noindent{{\textbf{DCE Threshold ($\epsilon$).} We evaluate the sensitivity to the DCE threshold $\epsilon$, which determines when to trigger selective refinement. Table~\ref{tab:epsilon} demonstrates that the DCE threshold exerts a relatively modest influence on overall inpainting quality. We find $\epsilon=2.5$ achieves the best trade-off between quality and efficiency.}}

\noindent {\textbf{Ablation on Alternatives of DCE.} To further justify our choice of Distributional Cross-Entropy (DCE) for the dynamic selective refinement mechanism, we conduct an ablation study to compare DCE against LPIPS, a widely used perceptual metric. The quantitative results are reported in Table~\ref{tab:dce_ablation}. As shown, our proposed DCE metric demonstrates better performance, achieving a significantly lower FID of 39.94 compared to LPIPS (58.79). While LPIPS is a robust perceptual metric for clean RGB images, it struggles to provide accurate structural evaluations on noisy intermediate states. In contrast, our DCE metric, which directly computes the cross-entropy of histogram distributions, is robust to spatial misalignment and less sensitive to the uniform diffusion noise. This enables our mechanism to accurately evaluate the distributional harmony and effectively trigger the refinement process.}
\begin{table}[h]
\small
\centering
\caption{{Sensitivity analysis of DCE threshold $\epsilon$ on ImageNet (Half masks).}}
\label{tab:epsilon}
\begin{tabular}{c|cc}
\hline
{$\epsilon$} & {LPIPS$\downarrow$} & {FID$\downarrow$} \\
\hline
{1.0} & {0.327} & {39.73} \\
{1.5} & {0.327} & {39.79} \\
{2.0} & {0.328} & {39.85} \\
{2.5} & {0.328}& {39.94} \\
{3.0} & {0.334}& {41.04} \\
\hline
\end{tabular}
\end{table}

\begin{table}[h]
\small
\centering
\caption{{Ablation on alternatives of DCE.}}
\label{tab:dce_ablation}
\begin{tabular}{c|cc}
\hline
{$Configuration$}& {LPIPS$\downarrow$} & {FID$\downarrow$} \\
\hline

{LPIPS}& {0.342}& {54.70}\\
{DCE}& {0.328}& {39.94}\\
\hline
\end{tabular}
\end{table}
\section{Conclusion}
\label{Section 5}
In this paper, we propose the Initial Seed refined Diffusion Model (IS-Diff) to set a promising direction for the diffusion process with a refined initial seed. 
Estimated data distribution from the unmasked images is extracted as a primary initial seed. {Furthermore, an iterative refinement mechanism is introduced to seek finer initialization by adjusting the initialization-to-noise ratio dynamically.}  
Extensive experimental results demonstrate that our method can generate more harmonious and realistic results in a training-free, efficient, and flexible manner.\\
\begin{figure}
    \centering
    \includegraphics[width=\linewidth]{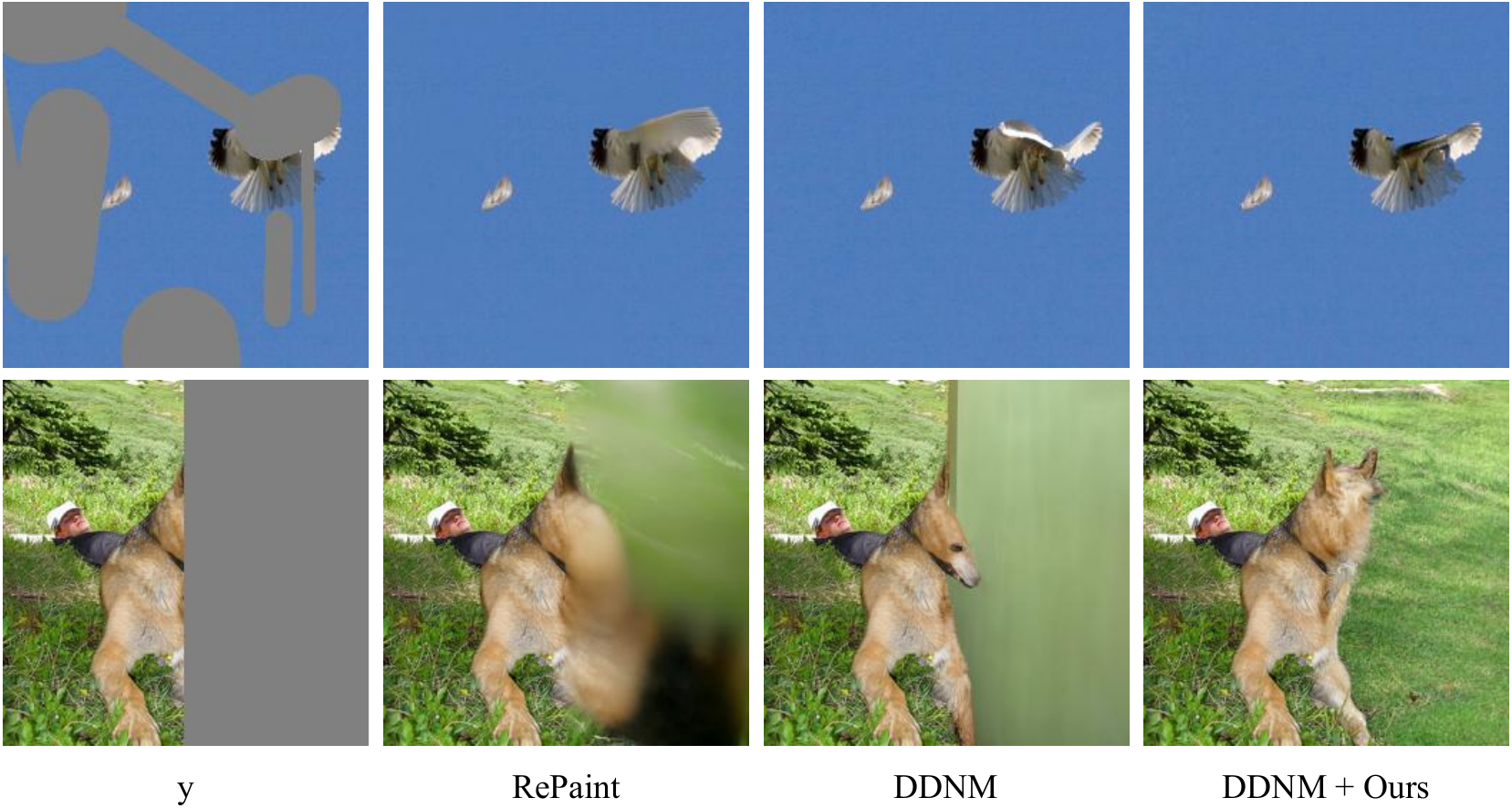}
    \caption{Failure Cases. Our method and other baselines usually fail to generate reasonable content when facing marginal occlusion of the foreground objects. }
    \label{fig:failure}
\end{figure}
\textbf{Limitations and Failure Cases.} While our IS-Diff significantly improves the robustness and realism of diffusion-based inpainting, it still faces challenges in extreme occlusion scenarios. Specifically, when the contextual clues of a specific foreground object are explicit but exceedingly weak (e.g., when over 95\% of the object is masked), our method may struggle to reconstruct the missing semantics accurately. This failure arises because the severely truncated visible portion provides insufficient visual clue for inpainting process. It is worth noting that other state-of-the-art diffusion baselines similarly fail to generate reasonable content under such extreme conditions due to the severe absence of semantic anchors. As shown in Fig.~\ref{fig:failure}, all methods fail to generate legs of the man in the bottom row, but our method still generates consistent textures in most regions. Furthermore, as with previous diffusion-based pipelines, our method remains computationally slower than GAN-based and autoregressive-based approaches. \\
\textbf{Social Impact.} Our proposed IS-Diff, as an image inpainting method, can be used for image editing, object removal, and other typical applications. Moreover, IS-Diff can be easily integrated into any diffusion-based methods to enhance their inpainting capabilities, which will further boost related researches. However, our method can be abused for image manipulation potentially. 

\section*{Acknowledgment}
This work was partially supported by the Key Research and Development Program of Yunnan Province under Grant 202403AA080002, and partially supported by the National Natural Science Foundation of China under Grant 62471344 and 62225113.
{
    \bibliographystyle{IEEEtran}
    \bibliography{main}
}

\newpage
\begin{IEEEbiography}[{\includegraphics[
width=1in,height=1.25in,clip,keepaspectratio]
{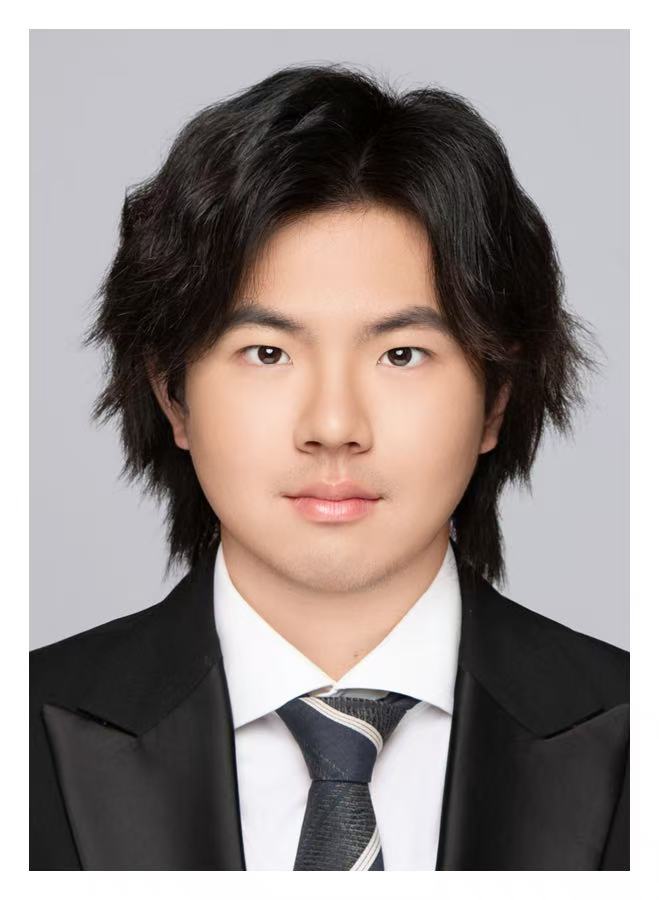}}]{Yongzhe Lyu}
received the B.E. degree from Wuhan University, Wuhan, China, in 2026.
His research interests include image restoration and 4D reconstruction.
\end{IEEEbiography}
\begin{IEEEbiography}
[{\includegraphics[width=1in,height=1.25in,clip,keepaspectratio]
{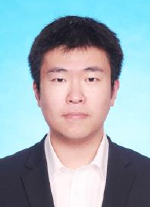}}]{Yu Wu}
is a Professor with the School of Computer Science at Wuhan University,
China. He received his Ph.D. degree from the University of Technology
Sydney, Australia, in 2021. From 2021 to 2022, he was a postdoctoral
researcher at Princeton University. His research interests include
controllable generation and multimodal perception. He received the
AAAI New Faculty Highlight in 2024 and the Google Ph.D. Fellowship in
2020. He has served as an Area Chair for CVPR, ICCV, ECCV, and NeurIPS,
and as the Workshop Chair for CVPR 2023.
\end{IEEEbiography}
\begin{IEEEbiography}[{\includegraphics[width=1in,height=1.25in,clip,keepaspectratio]{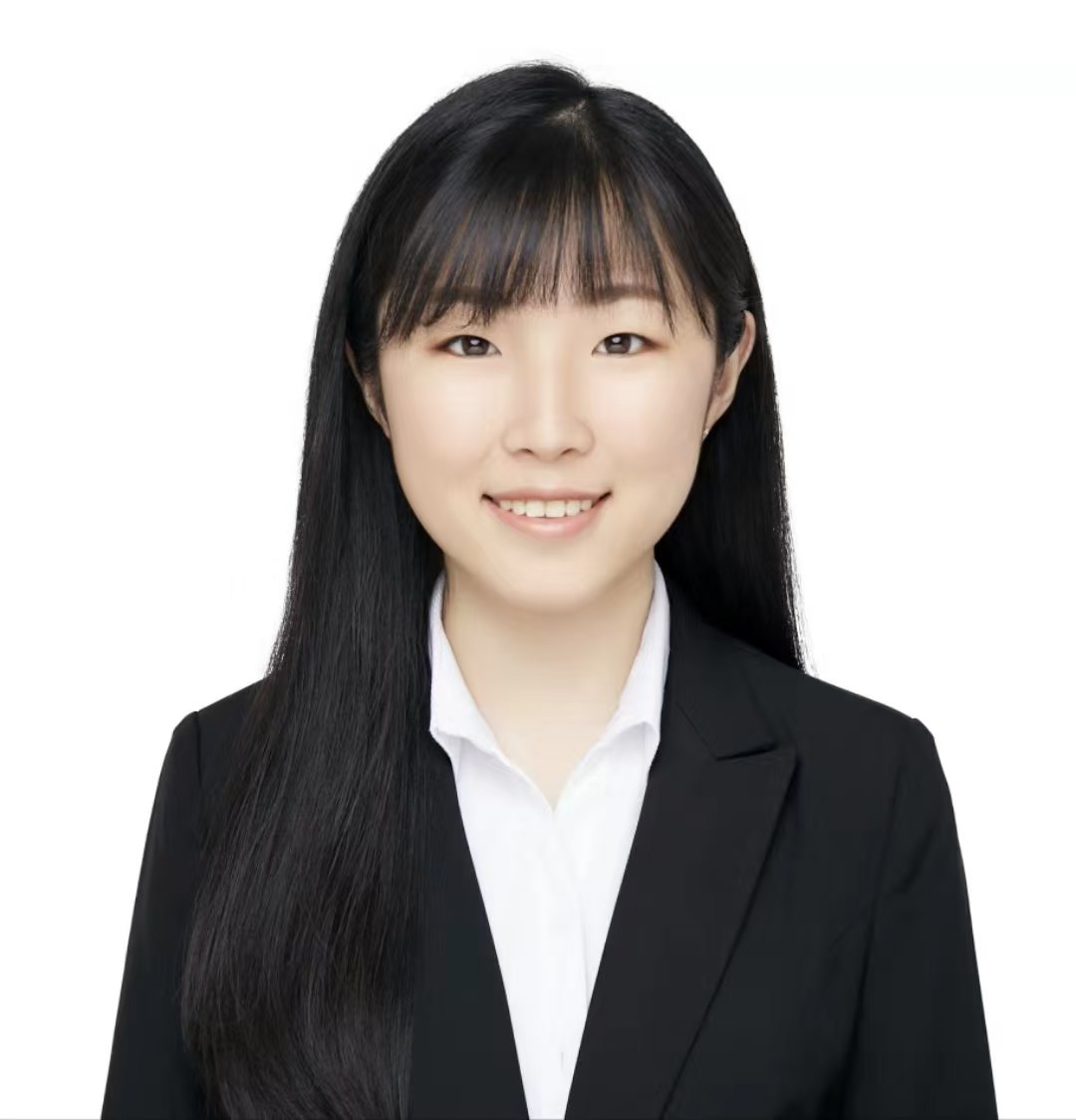}}]{Yutian Lin} received the B.E. degree from Zhejiang University, China, in 2016, and the Ph.D. degree from the Center for Artificial Intelligence, University of Technology Sydney, Australia, in 2019. She is currently an associate professor in Wuhan University, China. Her research interests include person re-ID and related applications, unsupervised learning and self-supervised learning.  
\end{IEEEbiography}
\begin{IEEEbiography}[{\includegraphics[width=1in,height=1.25in,clip]{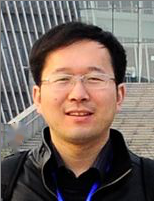}}]{Bo Du} (Senior Member, IEEE) received the Ph.D. degree in photogrammetry and remote sensing from the State Key Lab of Information Engineering in
Surveying, Mapping and Remote Sensing, Wuhan University, Wuhan, China, in 2010. He is currently a Professor with National Engineering Research Center for Multimedia Software, Institute of Artificial Intelligence, Wuhan University, Wuhan, China. His major research interests include pattern recognition, hyperspectral image processing, machine learning, and signal processing.
\end{IEEEbiography}

\newpage

\end{document}